\pdfoutput=1
\documentclass[11pt]{article}

\usepackage[final]{acl}
\usepackage{times}
\usepackage{latexsym}
\usepackage{pifont}
\usepackage{booktabs}
\usepackage{xcolor}
\usepackage{array}
\usepackage{multirow}
\usepackage{inconsolata}
\usepackage{amsmath,graphicx}
\usepackage{tabularx}
\usepackage{amssymb}
\usepackage{amsthm}
\usepackage{cleveref}
\usepackage{pifont}
\usepackage{cancel}
\usepackage{lipsum}
\usepackage{fvextra}
\usepackage{booktabs}
\usepackage{algorithm}
\usepackage{enumitem}
\usepackage{ragged2e}
\usepackage{subcaption}
\usepackage{url}

\usepackage[T1]{fontenc}
\usepackage[utf8]{inputenc}

\usepackage{microtype}

\usepackage{inconsolata}

\usepackage{graphicx}
\usepackage{subcaption}
\usepackage{booktabs}
\usepackage{multirow}
\usepackage[most]{tcolorbox}
\usepackage{titling}
\usepackage{xcolor}
\usepackage{pifont}
\usepackage[dvipsnames]{xcolor}
\usepackage{algpseudocode}
\usepackage{setspace}

\newcommand{\gptfouro}{\textsc{GPT-4o-mini}}
\newcommand{\wikitq}{\textsc{WiKiTQ}}
\newcommand{\tabfact}{\textsc{TabFact}}
\newcommand{\multiagent}{\textsc{MAPLE}}

\title{\includegraphics[height=1em]{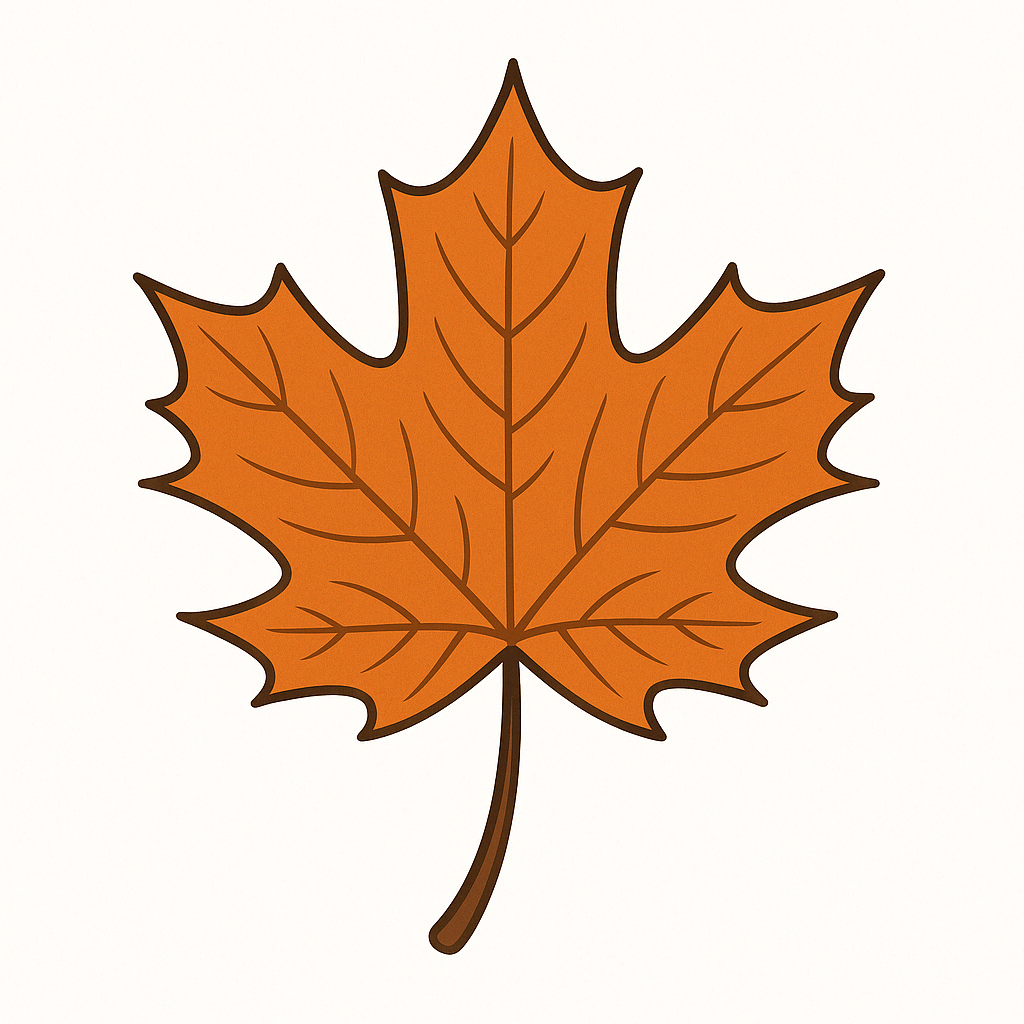} MAPLE: Multi-Agent Adaptive Planning with Long-Term Memory for Table Reasoning}

\author{Ye Bai\textsuperscript{1}, Minghan Wang\textsuperscript{1}, \textbf{Thuy-Trang Vu}\textsuperscript{1} \\
  \textsuperscript{1}Department of Data Science \& AI, Monash University \\
  \texttt{\{firstname.lastname\}}@monash.edu
}

\begin{document}
\maketitle

\begin{abstract}

Table-based question answering requires complex reasoning capabilities that current LLMs struggle to achieve with single-pass inference. Existing approaches, such as Chain-of-Thought reasoning and question decomposition, lack error detection mechanisms and discard problem-solving experiences, contrasting sharply with how humans tackle such problems. In this paper, we propose \multiagent~ (\textbf{M}ulti-agent \textbf{A}daptive \textbf{P}lanning with \textbf{L}ong-term m\textbf{E}mory), a novel framework that mimics human problem-solving through specialized cognitive agents working in a feedback-driven loop. \multiagent~ integrates 4 key components: (1) a Solver using the ReAct paradigm for reasoning, (2) a Checker for answer verification, (3) a Reflector for error diagnosis and strategy correction, and (4) an Archiver managing long-term memory for experience reuse and evolution. Experiments on \wikitq~and \textsc{TabFact} demonstrate significant improvements over existing methods, achieving state-of-the-art performance across multiple LLM backbones.\footnote{Dataset and code are available at \url{https://github.com/bettyandv/MAPLE-table-reasoning}}
%Ablation studies confirm that each component substantially contributes to the framework's effectiveness. Our memory analysis further reveals that logical reasoning errors and numerical operation failures account for most remaining challenges, providing valuable insights for future table reasoning research.

\end{abstract}
\section{Introduction}

Tables represent one of the most prevalent forms of semi-structured data, organizing information systematically across domains ranging from scientific research to business analytics~\citep{10.1145/3626772.3661384}. However, answering questions over tables presents unique challenges, requiring multi-step reasoning over structured data, recognition of implicit relationships between cells, and precise contextual interpretation~\citep{Lu_2025}. These challenges make table-based question answering (QA) particularly difficult for Large Language Models (LLMs), as they must navigate tabular data structure while performing sophisticated reasoning to derive accurate answers, capabilities that current LLMs struggle to achieve with single-pass inference.

Existing table reasoning frameworks exhibit several limitations. Single-forward-pass methods~\citep{cheng_binding_2023,ye_large_2023} lack error detection mechanisms, allowing mistakes to propagate through solutions. ReAct-based approaches~\citep{wang_chain--table_2024,zhang_reactable_2023} provide environmental feedback but lack systematic verification. On the other hand, multi-agent approaches primarily focus on output refinement rather than comprehensive reasoning improvement~\citep{ye_large_2023,yu2025tablecriticmultiagentframeworkcollaborative}. Additionally, current systems discard problem-solving experiences after completion, preventing transferable knowledge accumulation across tasks. It contrasts with human problem-solving: \textbf{when tackling complex tabular problems, humans methodically work through solutions, verify results, reflect on mistakes, and accumulate experiences for future strategies.}

\begin{figure*}[t]
    \centering
    \includegraphics[width=1\textwidth]{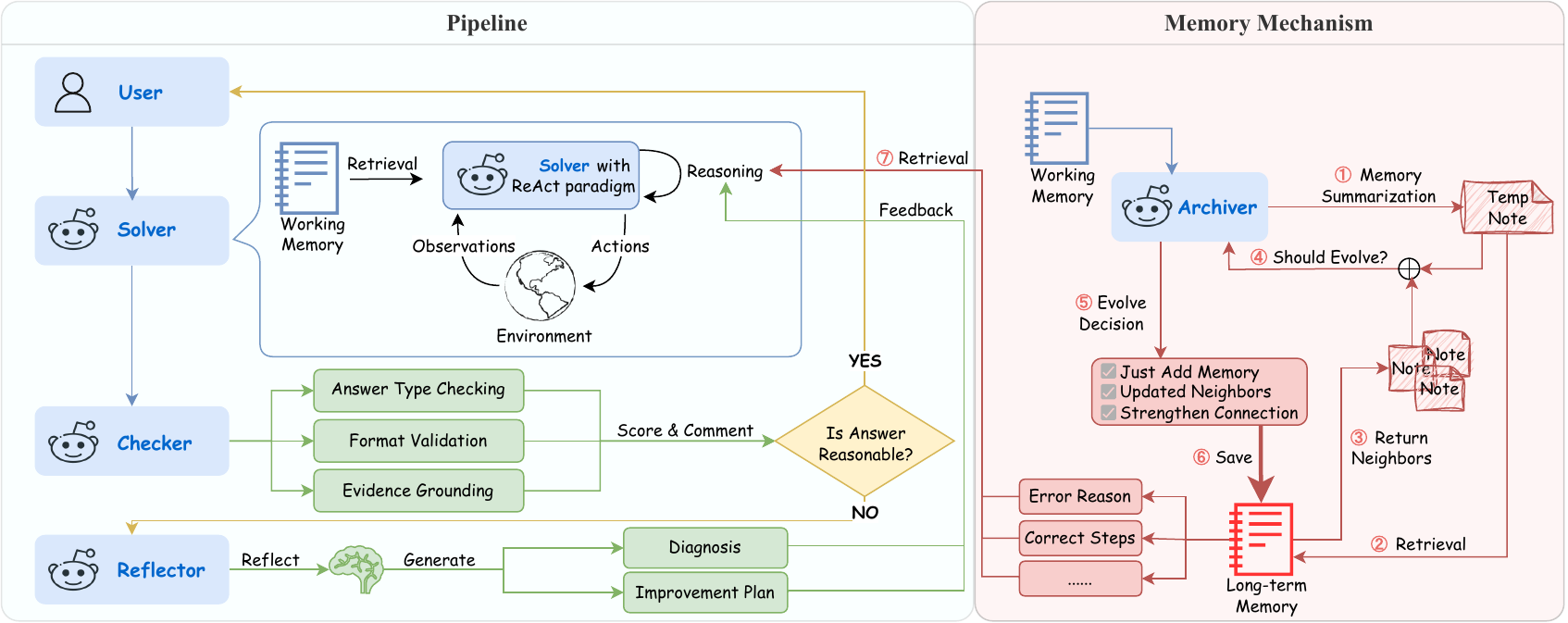}
    \caption{
    % The overall pipeline of \multiagent. The framework coordinates 4 agents---Solver, Checker, Reflector, and Advisor---to collaboratively solve table-based QA tasks. The Solver conducts iterative reasoning under the ReAct paradigm, the Checker evaluates answer quality, the Reflector analyzes reasoning errors, and the Advisor manages and evolves long-term memory.
    The \multiagent~ framework pipeline. 4 agents work collaboratively in a feedback loop: the Solver conducts iterative reasoning using ReAct, the Checker evaluates answer quality, the Reflector diagnoses errors and suggests improvements, and the Archiver manages an evolving long-term memory. This architecture enables dynamic adaptation both within tasks and across similar problems, mirroring human cognitive problem-solving processes.}
    % \vspace{-1em}
    \label{fig:ModelFramework}
\end{figure*}

To address these limitations, we propose \multiagent~ (\textbf{M}ulti-agent \textbf{A}daptive \textbf{P}lanning with \textbf{L}ong-term m\textbf{E}mory), a novel framework that mimics human problem-solving through specialized cognitive agents in a feedback-driven loop.  As illustrated in~\Cref{fig:ModelFramework}, \multiagent~ decomposes reasoning into distinct stages: reasoning, verification, reflection, and memory evolution, each managed by a dedicated agent. Our framework implements a feedback-driven cycle with the Solver conducting iterative reasoning, the Checker performing quality assessment, the Reflector diagnosing errors and suggesting improvements, and the Archiver managing long-term memory for cross-task learning. This architecture enables dynamic adaptation both within tasks and across similar problems, mirroring human cognitive processes.% in ways existing systems cannot replicate. 

% Experiments on WikiTQ and TabFact demonstrate that MAPLE significantly outperforms existing methods, achieving state-of-the-art performance across multiple LLM backbones. Ablation studies confirm that each component substantially contributes to the framework's effectiveness. Our memory analysis further reveals that logical reasoning errors and numerical operation failures account for nearly 80\% of remaining challenges, providing valuable insights for future table reasoning research. This analysis demonstrates that our memory system serves not only as a performance enhancer but also as a diagnostic tool, identifying specific reasoning bottlenecks that future work should prioritize.

Experiments on \wikitq~ and \textsc{TabFact} demonstrate that \multiagent~ significantly outperforms existing methods across multiple LLM backbones. Ablation studies confirm each component substantially contributes to the framework's effectiveness. Our memory analysis reveals that logical reasoning errors and numerical operation failures account for nearly 80\% of remaining challenges, providing valuable insights for future research and serving as both a performance enhancer and diagnostic tool.

Our contributions include: \textbf{(1)} a multi-agent framework implementing adaptive planning through feedback-driven reasoning; \textbf{(2)} a specialized verification and reflection mechanism providing targeted diagnostic feedback; \textbf{(3)} a structured long-term memory system that distills experiences and categorizes errors; and \textbf{(4)} state-of-the-art performance on \wikitq~ and \textsc{TabFact} benchmarks. These innovations address fundamental limitations in current approaches, creating a system that mirrors human cognitive processes while improving performance on complex table reasoning tasks.

\section{Related Work}

% short version：
\paragraph{Multi-agent System}
Multi-agent systems (MAS) have gained attention for leveraging collective intelligence in complex tasks. Current applications span problem-solving domains such as software development~\citep{li2023camel,du2023improvingfactualityreasoninglanguage,qian2024chatdevcommunicativeagentssoftware,huang2024agentcodermultiagentbasedcodegeneration}, embodied robotic coordination~\citep{dasgupta2023collaboratinglanguagemodelsembodied,mandi2023rocodialecticmultirobotcollaboration,zhang2024buildingcooperativeembodiedagents,yu2025conavgptmultirobotcooperativevisual}, and scientific Debate~\citep{du2023improvingfactualityreasoninglanguage,Xiong_2023,tang2024medagentslargelanguagemodels}. World simulation applications include social behavior modeling~\citep{park2022socialsimulacracreatingpopulated,gao2023s3socialnetworksimulationlarge,chen2025multiagentconsensusseekinglarge}, policy simulation~\citep{xiao2023simulatingpublicadministrationcrisis,hua2024warpeacewaragentlarge}, economic forecasting~\citep{horton2023largelanguagemodelssimulated,li2023tradinggptmultiagentlayeredmemory,zhao2024competeaiunderstandingcompetitiondynamics}, and gaming~\citep{light2023avalonbenchevaluatingllmsplaying,wang2023avalonsgamethoughtsbattle,xu2024languageagentsreinforcementlearning}. Several MAS frameworks, including MetaGPT~\citep{hong2024metagptmetaprogrammingmultiagent}, CAMEL~\citep{li2023camel}, and AutoGen~\citep{wu2023autogenenablingnextgenllm}, have emerged to facilitate the implementation of these systems.

% In table reasoning, MAS applications remain limited. Existing frameworks like Dater~\citep{ye_large_2023} employ minimal agent structures tailored to question-answering tasks without extensive collaboration. Similarly, Table-Critic~\citep{yu2025tablecriticmultiagentframeworkcollaborative} primarily focuses on output refinement rather than interactive reasoning.

\paragraph{Memory Mechanisim for Agent}
Memory enables agents to perform coherent, context-aware reasoning across extended tasks. In multi-agent systems, memory serves as the cognitive foundation for retaining observations, decisions, and interaction histories—critical elements for consistent collaboration and adaptation over time~\citep{sumers2024cognitivearchitectureslanguageagents}.

Memory formats determine how content is stored and utilized, with several prominent approaches emerging in recent work. Natural language representations offer semantic richness and interpretability, as demonstrated in Reflexion~\citep{shinn2023reflexionlanguageagentsverbal}, Voyager~\citep{wang2023voyageropenendedembodiedagent}, and Generative Agents~\citep{park2023generativeagentsinteractivesimulacra}. Vector embeddings enable efficient similarity-based retrieval, a technique central to systems like MemoryBank~\citep{zhong2023memorybankenhancinglargelanguage}, A-MEM~\citep{xu2025amemagenticmemoryllm}, and ChatDev~\citep{qian2024chatdevcommunicativeagentssoftware}. Meanwhile, structured formats support symbolic reasoning and precise queries, approaches adopted by ChatDB~\citep{hu2023chatdbaugmentingllmsdatabases} and DB-GPT~\citep{zhou2023llmdba}. Beyond storage formats, recent advances have introduced innovative management strategies, including complete interaction storage~\citep{zhong2023memorybankenhancinglargelanguage,modarressi2024retllmgeneralreadwritememory}, cache-like designs~\citep{packer2024memgptllmsoperatingsystems}, and controller-based architectures~\citep{wang2025scmenhancinglargelanguage} that dynamically prioritize and maintain relevant information during extended reasoning processes.

Despite these advances, few existing systems integrate all these memory dimensions within a coherent architecture specifically designed for complex reasoning tasks like table-based QA, where both structured knowledge and flexible retrieval are essential for effective performance.

\paragraph{Table Reasoning} Research on table reasoning can be broadly classified into fine-tuning-based and prompting-based methods. Fine-tuning approaches like TAPAS~\citep{herzig_tapas_2020}, Pasta~\citep{gu_pasta_2022}, TUTA~\citep{wang_tuta_2021}, and TAPEX~\citep{liu_tapex_2022} adapt pre-trained language models to encode table semantics through specialized training objectives. Other works focus on improving alignment between natural language queries and structured data~\citep{eisenschlos-etal-2020-understanding, jiang-etal-2022-omnitab}. Despite their effectiveness, these methods typically require extensive annotated data and feature static reasoning processes without adaptive correction mechanisms.

Prompting-based methods leverage LLMs with minimal training data requirements. Techniques like Chain-of-Thought~\citep{wei2023chainofthoughtpromptingelicitsreasoning}, Least-to-Most~\citep{zhou2023leasttomostpromptingenablescomplex} and Dater~\citep{ye_large_2023} perform reasoning by decomposing tasks into explicit steps. Table-specific adaptations include Binder~\citep{cheng_binding_2023}, Chain-of-Table~\citep{wang_chain--table_2024}, ReAcTable~\citep{zhang_reactable_2023}, and Table-Critic~\citep{yu2025tablecriticmultiagentframeworkcollaborative}, incorporating agent collaboration or ReAct-style reasoning.

Existing table reasoning methods exhibit significant limitations across key design dimensions. While some implement multi-agent architectures or ReAct-based reasoning, none integrates all critical components: dynamic planning, reflection mechanisms, self-refinement, and long-term memory. A detailed comparison of these methods is provided in~\Cref{related_work_appendix}. Our proposed \multiagent~ framework addresses these gaps by combining collaborative verification, adaptive planning, and evolving memory structures to achieve more robust and accurate table reasoning.

\section{\multiagent~Framework}
\subsection{Overview}

\begin{table*}[t]
\centering
\resizebox{\textwidth}{!}{%
\begin{tabular}{@{}llll@{}}
\toprule
Agent & Input & Output & Function \\ \midrule
Solver & $\mathcal{T}$ or $t', q, \tau, r, \mathcal{M}$ & $t'$ or $a_s$ & Progressive reasoning with real-time environmental feedback \\
Checker & $\mathcal{T}, q, a_s$ & $\mathcal{F}$ & 
Verifies answer type, format and evidence grounding \\
Reflector & $\mathcal{T}, q, \tau, a_s, \mathcal{F}$ & $d, p$ & Diagnoses errors and generates targeted improvement plans \\
\multirow{3}{*}{Archiver} & $\mathcal{T}, q, a_m, a_g, \tau, d, p$ & $m$ & Distills experiences into structured memory notes \\
 & $\mathcal{T}, q, m, \mathcal{M}_l, k, \delta$ & $\mathcal{N}$ & 
 Retrieves contextually relevant experiences for current tasks
 % Retrieves relevant memory for reasoning / evolution
 \\
 & $m, \mathcal{N}$ & $e$ & Evolves memory through semantic clustering and connection \\ \bottomrule
\end{tabular}%
}
\caption{Overview of specialized agents in \multiagent. Each agent performs distinct cognitive functions with specific input-output patterns. This modular design allows for verification, reflection, and experience reuse across tasks.
% Overview of the core components in our framework. Each agent operates over a specific subset of inputs and outputs, serving distinct roles in reasoning, verification, reflection, memory management, and retrieval. 
% Each agent is responsible for a distinct phase of the table-based QA process, including progressive reasoning, answer verification, error reflection, and long-term memory management. Inputs and outputs for each agent are listed alongside their primary responsibilities.
}
\vspace{-0.8em}
\label{tab:agent_summary}
\end{table*}

Inspired by human problem-solving processes, we propose \multiagent~ (\textbf{M}ulti-agent \textbf{A}daptive \textbf{P}lanning with \textbf{L}ong-term m\textbf{E}mory), a novel framework that addresses a critical limitation in existing systems: their inability to adapt, reflect, and learn from experience. As illustrated in \Cref{fig:ModelFramework} and formalized in \Cref{alg:overall_pipeline}, \multiagent~ decomposes reasoning into four distinct stages, each managed by a dedicated agent that fulfills a specialized cognitive function.

The Solver conducts progressive reasoning using the ReAct paradigm, enabling dynamic interaction with the table environment. The Checker performs multi-dimensional verification across answer type, format, and evidence grounding. The Reflector diagnoses reasoning errors and generates targeted improvement plans when verification fails. Finally, the Archiver manages long-term memory, facilitating experience reuse across similar problems. To more concretely illustrate the flow of our framework, we present a representative case study in Appendix~\ref{appendix_case_study}.

A key innovation in \multiagent~ is its feedback-driven, multi-round reasoning cycle that enables continuous refinement of reasoning strategies through deliberate planning and adaptation. This allows agents to correct errors and improve solutions through multiple attempts, \textbf{mimicking how humans iteratively improve their problem-solving approaches.}

Additionally, while existing systems discard reasoning experiences after completion, our approach implements selective integration and strategic evolution of memory. The system filters redundant experiences, distills valuable problem-solving patterns into structured notes, and evolves the memory base through semantic clustering. This integration of adaptive planning with evolving memory enables \multiagent~ to leverage past experiences, avoid repeated errors, and continuously improve reasoning capabilities across similar problem types.

\subsection{Agent Roles}
Table~\ref{tab:agent_summary} summarizes the specialized responsibilities and input-output specifications of each agent in our framework. Below, we formally define how these agents interact within the \multiagent~ architecture to create a cohesive reasoning system that surpasses traditional single-pass approaches.

\subsubsection{Solver} 

The Solver agent $(\mathcal{S})$ leverages the \textbf{ReAct} paradigm~\citep{yao2023reactsynergizingreasoningacting} to establish genuine environmental interaction with tabular data. After each table operation, the Solver reassesses the environment to determine whether to perform additional manipulations or derive an answer. This iterative reasoning-acting cycle enables strategic adaptation based on real-time feedback from the manipulated table state.

Formally, the Solver operates according to the following process, given input $(I, q, \tau, r, \mathcal{M})$:
\begin{equation}
  \label{eq:solver}
\lambda = \pi_\text{solver}(I, q, \tau, r, \mathcal{M})
% \begin{cases}
%     t', & \lambda_I = \text{operate table}; \\
%     a_s,  & \lambda_I = \text{derive answer}.
% \end{cases}
\end{equation}

where $\lambda \in \{t', a_s\}$, which can either be the operated table $t'$ (if the reasoning is not done yet) or the final answer ($a_s$) of the Solver, \(I \in \{\mathcal{T}, t' \}\) represents the current environment (original table $\mathcal{T}$ or intermediate table \(t'\)), \(q\) is the question, \(\tau\) denotes previous Solver operation history, and \(r\) is the remaining attempts.

A distinctive feature of our approach is the integration of dual memory systems. The memory input is defined as $\mathcal{M} = \{\mathcal{M}_w, \mathcal{M}_l\}$, where working memory $\mathcal{M}_w$ contains Reflector feedback ($d,p$) providing diagnostic insights when prior attempts failed. Long-term memory $\mathcal{M}_l$ retrieves relevant historical experiences ($\mathcal{N}_{\text{solver}}$), including similar questions, proven strategies and common pitfalls, creating a knowledge repository that enriches the reasoning process.

After each interaction, the updated environment \(I\) is fed back to the Solver, enabling continuous adaptation based on the evolving table state. Through this feedback loop, the Solver can progressively refine its understanding and approach until reaching a satisfactory answer.
The complete prompt is provided in Appendix~\ref{sec:prompt_examples}.

\subsubsection{Checker}

The Checker agent $(\mathcal{C})$ introduces a critical verification layer based on structured feedback principles. 
% Drawing inspiration from the FEEDBACK module in SELF-REFINE \citep{madaan2023selfrefineiterativerefinementselffeedback}, which provides multi-faceted feedback for dialogue responses, we adapt and extend these qualitative aspects specifically for table reasoning tasks. 
Given table $\mathcal{T}$, question \(q\), and Solver's answer \(a_s\), the Checker evaluates the output according to three essential criteria:

\begin{enumerate}[noitemsep,topsep=2pt,parsep=2pt,partopsep=2pt]
\item{\textbf{Answer Type Checking:}} Evaluates whether the answer matches the expected type implied by the question. For instance, if a question asks "How many medals did the country win?", a numerical value like "5" is expected, not a country name like "USA".
\item{\textbf{Format Validation:}} Assesses conformity to prescribed formatting rules. If the expected output is a single numerical value ("24"), but the answer includes calculation steps ("4 × (1 + 2 + 3) = 24"), this violates format requirements that mandate only the final result.
\item{\textbf{Evidence Grounding:}} Verifies that the answer is properly supported by evidence in the table data. If a question asks "Which country won the most gold medals?" and the table only lists "USA", "China", and "Japan", an answer of "Germany" would violate evidence grounding as it does not appear in the table.
\end{enumerate}

For each criterion $i \in \{\text{type}, \text{format}, \text{evidence}\}$, the Checker assigns a score $s_i \in \{0, 1, 2\}$ with an explanatory comment $c_i$, where 0 indicates the requirement is not met, 1 indicates it is partially met, and 2 indicates it is fully met or not applicable. Formally, this evaluation process is defined as:

\begin{equation}
  \label{eq:checker}
  \{(s_i, c_i)\}_{i=1}^3 = \pi_\text{checker}(\mathcal{T}, q, a_s)
\end{equation}

where $\pi_\text{checker}$ represents the evaluation function mapping the input triplet to aspect-specific scores and comments. This structured feedback $\mathcal{F}$ includes a total score $s_{\text{total}} = \sum_{i=1}^3 s_i$ and an aggregated summary, enabling the Reflector agent to diagnose errors and generate improvement strategies.

The Checker forms an integral component of the reasoning cycle, creating a feedback loop that drives continuous improvement. By systematically evaluating answers across multiple dimensions, it helps identify specific weaknesses in the reasoning process rather than merely flagging incorrect answers. The complete prompt is provided in Appendix~\ref{sec:prompt_examples}.

\subsubsection{Reflector}
The Reflector agent $(\mathcal{R})$ implements a metacognitive capability essential for advanced reasoning systems, analyzing failures, diagnosing root causes, and generating strategic corrections. This agent drives continuous improvement through deliberate adaptation, providing the critical link between error detection and strategy refinement.

Given the table $\mathcal{T}$, question \(q\), Solver's reasoning trace $\tau$, Solver's answer \(a_s\), and Checker feedback $\mathcal{F}$, the Reflector analyzes reasoning deficiencies and formulates targeted remediation strategies:

\begin{equation}
  \label{eq:reflector}
  (d, p) = \pi_{\text{reflector}}(\mathcal{T}, q, \tau, a_s, \mathcal{F})
\end{equation}

where $d$ represents a concise diagnostic summary identifying critical reasoning errors, and $p$ outlines an actionable improvement plan with step-by-step corrections for subsequent attempts.

This reflection mechanism creates a powerful feedback loop: the Solver adapts based on precise diagnosis and targeted suggestions, rather than blindly attempting alternatives. Without such directed feedback, traditional systems often repeat the same errors, unable to identify reasoning flaws. The Reflector examines not just answer correctness, but the entire reasoning trajectory, pinpointing where logical connections faltered, operations were misapplied, or question intent was misunderstood.

By implementing this metacognitive layer, \multiagent~ progressively refines its reasoning strategies across multiple attempts, achieving within-task learning that static reasoning systems cannot replicate. The  prompt template detailed in Appendix~\ref{sec:prompt_examples}.

% {\small
\begin{algorithm*}[t]
\caption{:~\multiagent's Adaptive Reasoning Loop with Multi-Agent Feedback}
\label{alg:overall_pipeline}
% \begingroup
% \setstretch{0.95}
\begin{algorithmic}[1]
\State \textbf{Input:} Table $\mathcal{T}$, Question $q$, Memory $\mathcal{M} \in \{\mathcal{M}_w, \mathcal{M}_l\}$, Remaining Attempts $r$, Neighbor Limit $k$, Similarity Threshold $\delta$
\State \textbf{Output:} Final Answer $a_m$
\State $Finished \gets \text{False}$ \Comment{Initialization flag}
\State $\tau \gets []$ \Comment{Initialize Solver operation history}
\While{$\neg Finished$ and $r > 0$}
    \State $ \mathcal{N}_{\text{solver}} \gets  \pi_{\text{mem-retrieve}}^{\text{solver}}(\mathcal{T}, q\mid \mathcal{M}_l, k, \delta)$  \Comment{Retrieve neighbor memories, Eq.~\ref{eq:advisor_re_solver}}
    \State $t'$ or $a_s \gets \pi_{\text{solver}}(I, q, \tau, r, \mathcal{M}
    % ,\mathcal{N}_{\text{solver}},d, p
    )$ \Comment{Operated table or answer of Solver, Eq.~\ref{eq:solver}} 
    \State $\tau.\text{append(LogSolverOperation($\mathcal{M}_w))$}$ \Comment{Log current operation in operation history}
    \If{$a_s \neq \texttt{<NOT\_READY>}$}
        \State $\mathcal{F} \gets \pi_{\text{checker}}(\mathcal{T}, q, a_s)$ \Comment{Evaluate answer, Eq.~\ref{eq:checker}}
        \If{$\mathcal{F}.\text{total\_score} == \texttt{FULL\_SCORE}$}
            \State $a_m \gets a_s$ \Comment{Accept Solver answer}
            \State $Finished \gets \text{True}$
        \Else
            \State $(d, p) \gets \pi_{\text{reflector}}(\mathcal{T}, q, \tau, a_s, \mathcal{F})$ \Comment{Diagnose and provide fix plan, Eq.~\ref{eq:reflector}}
            \State $\mathcal{M}_w \gets \text{UpdateWorkingMemory}(d, p)$ \Comment{Inject feedback for adaptive refinement}
        \EndIf
    \Else
        \State Continue ReAct reasoning on updated table $t'$ \Comment{Continue ReAct loop}
    \EndIf
    \State $r \gets r - 1$ \Comment{Decrease remaining attempts}
\EndWhile
\State \textbf{return} $a_m$ \Comment{Output model prediction}
\end{algorithmic}
% \endgroup
\end{algorithm*}
% }

\subsubsection{Archiver}

% The Archiver agent $(\mathcal{A})$ introduces experiential learning through structured memory management. This agent implements knowledge retention, retrieval, and evolution mechanisms that enable continual improvement across reasoning tasks. The Archiver operates through three complementary modules:

The Archiver agent $(\mathcal{A})$ introduces experiential learning through structured memory management. This agent implements knowledge retention, retrieval, and evolution mechanisms that enable continual improvement across reasoning tasks, with detailed working memory and long-term memory mechanisms described in~\Cref{appendix_memory}. The Archiver operates through three complementary modules:

\paragraph{Memory Summarization.} 

Given the current table $\mathcal{T}$, question \(q\), model-predicted answer \(a_m\), ground truth \(a_g\), Solver's reasoning history $\tau$, and Reflector's outputs $(d, p)$, the Archiver distills this rich context into a compact, structured memory note:

\begin{equation}
  \label{eq:advisor_sum}
  m = \pi_{\text{archiver-sum}}(\mathcal{T}, q, a_m, a_g, \tau, d, p)
\end{equation}

This summarization extracts critical semantic features like keywords, tags, required operations, error types, correct and incorrect steps, creating an informative memory note $m$ that captures the problem-solving episode's essence. These notes are stored in long-term memory $\mathcal{M}_l$ to guide future reasoning.

\paragraph{Memory Retrieval.} 
The system implements dual retrieval modes for different contexts:

\begin{enumerate}[noitemsep,topsep=2pt,parsep=2pt,partopsep=2pt]
    \item \textbf{Solver-time retrieval}: During question answering, the system retrieves relevant memory notes based on the current table and question. The top-$k$ semantically similar results within threshold $\delta$ are returned for prompt augmentation:
    \begin{equation}
    \label{eq:advisor_re_solver}
        \mathcal{N}_{\text{solver}} = \pi_{\text{mem-retrieve}}^{\text{solver}}(\mathcal{T}, q\mid \mathcal{M}_l, k, \delta)
    \end{equation}

    \item \textbf{Archiver-time retrieval}: During memory management, the system identifies neighboring notes for new candidates to inform evolution decisions:
    \begin{equation}
    \label{eq:advisor_re_advisor}
        \mathcal{N}_{\text{archiver}} = \pi_{\text{mem-retrieve}}^{\text{archiver}}(m \mid \mathcal{M}_l, k, \delta)
    \end{equation}
\end{enumerate}

This dual architecture enables both experiential guidance during active reasoning and strategic memory refinement during maintenance, creating a dynamic knowledge ecosystem that continuously improves.

\paragraph{Memory Evolution.} 
Our memory system actively evolves. Given the newly created memory $m$ and a set of retrieved neighbor memories $\mathcal{N}_{\text{archiver}}$, the Archiver determines whether the memory base should evolve:

\begin{equation}
  \label{eq:advisor_evol}
  e = \pi_{\text{archiver-evo}}(m, \mathcal{N}_{\text{archiver}})
\end{equation}

where $e$ specifies evolution decisions, including whether to evolve and what specific actions (strengthening connections or updating memory metadata) to take. This process enhances semantic clustering of related experiences, enabling more contextually relevant knowledge retrieval in subsequent tasks.

Through this sophisticated memory management approach, \multiagent~ transcends traditional systems that restart reasoning from scratch on each task. Instead, our framework builds an evolving knowledge repository that improves performance across similar problems through continuous refinement.
For a more detailed algorithm describing this memory evolution process, please see~\Cref{app:memory_algorithm}.
% ~\Cref{alg:memory_pipeline} formalizes the Advisor's protocol for evaluating new experiences, filtering redundant information, and strategically evolving the memory base. This selective approach maintains a diverse, high-quality knowledge repository that enhances reasoning capabilities across problem domains. 
The complete Archiver prompt is provided in Appendix~\ref{sec:prompt_examples}.

Having defined the specialized roles and interactions of each agent in our framework, we now formalize the complete reasoning procedure that orchestrates their collaborative operation. ~\Cref{alg:overall_pipeline} presents the adaptive reasoning loop of \multiagent, illustrating how multiple agents coordinate through a feedback-driven cycle to progressively refine reasoning strategies. This algorithm demonstrates several key innovations absent in traditional approaches: (1) iterative refinement through verification and reflection, (2) dynamic adaptation based on structured feedback, and (3) integration of experiential knowledge from similar problems.
% Through this multi-agent collaboration, \multiagent~ achieves a level of reasoning adaptability and error resilience that fundamentally surpasses static, single-pass methods in complex table-based question answering tasks.

\section{Experiments}

\subsection{Experimental Setup}
\begin{table*}[t]
\centering
\resizebox{\textwidth}{!}{%
\begin{tabular}{@{}lcccc|cccc@{}}
\toprule
\textbf{Dataset} & \multicolumn{4}{c|}{\textbf{\wikitq}} & \multicolumn{4}{c}{\textbf{\tabfact}} \\ \midrule
\textbf{Model} & \textbf{\gptfouro} & \textbf{\textsc{LLaMA3.3-70B}} & \textbf{\textsc{Qwen2.5-72B}} & \textbf{Average} & \textbf{\gptfouro} & \textbf{\textsc{LLaMA3.3-70B}} & \textbf{\textsc{Qwen2.5-72B}} & \textbf{Average} \\ \midrule
End-to-End QA & 46.64 & 45.58 & 35.93 & 42.72 & 71.69 & 75.49 & 63.49 & 70.22 \\
Few-Shot QA & 57.16 & 58.54 & 37.66 & 51.12 & 70.16 & 74.73 & 67.49 & 70.79 \\
Chain-of-Thought & 62.96 & {\underline{65.75}} & 63.12 & 63.94 & 71.94 & 75.89 & 77.67 & 75.17 \\ \midrule
Binder~\citep{cheng_binding_2023} & 60.24 & 65.26 & 63.64 & 63.05 & 75.45 & 74.54 & 80.20 & 76.73 \\
Dater~\citep{ye_large_2023} & 61.13 & 62.18 & 61.94 & 61.75 & 75.39 & 81.12 & 78.46 & 78.32 \\
Chain-of-Table~\citep{wang_chain--table_2024} & {\underline{62.98}} & 64.80 & {\underline{67.87}} & 65.22 & {79.05} & {\underline{83.25}} & 81.96 & 81.42 \\
ReAcTable~\citep{zhang_reactable_2023} & 62.79 & 63.07 & 63.74 & 63.20 & \underline{79.18} & 81.49 & {\underline{82.06}} & 80.91 \\ \midrule
\multirow{2}{*}{\textbf{\multiagent~(Ours)}} & \textbf{67.13} & \textbf{74.01} & \textbf{73.39} & \textbf{71.51} & \textbf{81.62} & \textbf{90.66} & \textbf{86.02} & \textbf{86.10} \\
 & \textcolor{red}{↑ 4.15} & \textcolor{red}{↑ 8.26} & \textcolor{red}{↑ 5.52} & \textcolor{red}{↑ 6.29} & \textcolor{red}{↑ 2.44} & \textcolor{red}{↑ 7.41} & \textcolor{red}{↑ 3.96} & \textcolor{red}{↑ 4.68} \\ \bottomrule
\end{tabular}%
}
\caption{Table reasoning accuracy on \textsc{WikiTQ} and \textsc{TabFact} using \textsc{Qwen2.5-72B}, \textsc{LLaMA3.3-70B}, and \textsc{GPT-4o-mini}. 
% The table compares \multiagent~ with baseline prompting strategies and recent methods. 
\textbf{Bold} denotes the best performance and \underline{underline} denotes the second-best performance in each column. \textcolor{red}{Red arrows} indicate improvements over the strongest baseline.}
\vspace{-1em}
\label{tab:main_result_long}
\end{table*}

\paragraph{Datasets.} We evaluate our approach on two standard benchmarks: (1) WikiTableQuestions (\wikitq)~\citep{pasupat_compositional_2015}: A widely used benchmark dataset for studying question answering over structured tables. It contains 14,149 question-answer pairs for training and 4,344 for testing, collected from 421 Wikipedia tables. The questions require different levels of reasoning, and the answers can be single values, lists of values, or derived results that are not explicitly present in the table. 
(2) \tabfact~\citep{chen2020tabfactlargescaledatasettablebased}: A benchmark for fact verification over tabular data, consisting of natural language statements paired with tables from diverse domains. Each statement is labeled as either entailed ("yes") or refuted ("no") based on the table content. The test set includes 2,024 statements across 298 tables, requiring models to perform complex reasoning to verify factual accuracy.

\paragraph{Baselines.} We compare our multi-agent framework against three categories of baseline approaches: 
(1) \textbf{Standard Reasoning}, where the model directly generates answers from the table and question. This includes End-to-End QA, which outputs the answer in a single step. Few-Shot QA, which adds example (Table, Question, Answer) triplets to guide the model. Chain-of-Thought~\citep{wei2023chainofthoughtpromptingelicitsreasoning}, which encourages the model to explain its reasoning process before answering. (2) \textbf{Program-Based Reasoning}, which guide the model to produce executable code for answering. Binder~\citep{cheng_binding_2023} prompts the model to generate Python or SQL code. Dater~\citep{ye_large_2023} breaks down the question and table into smaller parts for easier processing.
(3) \textbf{ReAct-Based Reasoning}: This approach integrates reasoning and acting in an iterative process, using external tools to assist decision-making. Chain-of-Table~\citep{wang_chain--table_2024} dynamically constructs intermediate tables to support reasoning. ReAcTable~\citep{zhang_reactable_2023} follows this paradigm by integrating SQL and Python executions to generate intermediate results and refine reasoning steps.

\paragraph{Implementation Details.} We conduct our experiments using 3 state-of-the-art LLMs: 
\gptfouro\footnote{https://platform.openai.com/docs/models/gpt-4o-mini}, 
\textsc{LLaMA3.3-70B-Instruct}~\citep{grattafiori2024llama3herdmodels}\footnote{https://huggingface.co/meta-llama/Llama-3.3-70B-Instruct} and \textsc{Qwen2.5-72B-Instruct}~\citep{qwen2025qwen25technicalreport}\footnote{https://huggingface.co/Qwen/Qwen2.5-72B-Instruct}. All models run on two NVIDIA A100 GPUs. The tabular input is converted into markdown format before being passed to the LLMs. We use in-context prompting by including task-specific examples, which are provided in \Cref{sec:prompt_examples}. Default decoding parameters are used throughout. For all baseline methods, we follow their original settings to ensure optimal performance.

\paragraph{Metrics.} For \wikitq, we compute denotation accuracy by measuring whether the predicted answer matches the gold answer, regardless of surface form. For \tabfact, where the task is framed as binary classification (“yes” or “no”), we evaluate model predictions using exact string matching against the ground truth labels.

\subsection{Main Results}
Table~\ref{tab:main_result_long} presents the performance comparison on \textsc{WikiTQ} and \textsc{TabFact} across \textsc{LLaMA3.3-70B}, \gptfouro~ and \textsc{Qwen2.5-72B}. Our proposed method, \multiagent, consistently outperforms all baselines across both datasets and model backbones. On \textsc{WikiTQ}, \multiagent~ achieves 74.01\% and 73.39\% accuracy with respective models, while on \textsc{TabFact}, it reaches 90.66\% and 86.02\%. These results represent substantial gains of up to +8.26\% on \textsc{WikiTQ} and +7.41\% on \textsc{TabFact} over the strongest baseline methods.

Compared to recent specialized frameworks like Chain-of-Table and ReAcTable, \multiagent~ demonstrates consistent improvements across both datasets. The gains are particularly pronounced on \textsc{WikiTQ} (+5.52\% with \textsc{Qwen2.5-72B}), aligning with our framework's strength in handling compositional reasoning tasks that require progressive refinement. For \textsc{TabFact}, the improvements confirm that our approach remains effective even in binary classification settings. 
% Notably, while baseline methods show varying performance between model backbones, \multiagent~ maintains its superiority regardless of the underlying LLM, suggesting that our architecture provides fundamental reasoning advantages independent of specific base model capabilities. 
For analysis of how table size affects reasoning performance and the impact of multi-round reasoning, see~\Cref{appendix_additional_exp}.

These results validate the core design principles of \multiagent: dynamic adaptive planning through multi-round feedback loops, specialized agent roles with distinct cognitive functions, and progressive knowledge accumulation through long-term memory. 
% Together, these mechanisms create a reasoning system that significantly enhances LLMs' ability to perform complex tabular reasoning tasks.

\subsection{Evaluating Agent Contributions}
To assess the contribution of each agent in our framework, we conduct an ablation study using the LLaMA3.3-70B model on \wikitq. As shown in Table~\ref{tab:ablation}, we incrementally introduce agents into the \multiagent~ pipeline. Starting from a baseline that directly predicts answers without reasoning, we observe that adding the Solver agent alone leads to a substantial performance boost of +18.2 points, confirming the effectiveness of our ReAct-style multi-step reasoning. Incorporating the Checker further improves accuracy (+2.1), suggesting that verifying answer quality plays a crucial role in reducing erroneous outputs. Introducing the Reflector yields an additional significant gain (+5.2), highlighting the importance of iterative reflection and error correction. Finally, equipping the system with the Archiver enables long-term memory utilization, resulting in peak accuracy of 74.01. 
% These findings demonstrate the complementary roles of all four agents and validate the cumulative benefit of \multiagent's modular design.
\begin{table}[t]
\centering
\resizebox{\columnwidth}{!}{%
\begin{tabular}{@{}lc@{}}
\toprule
\textbf{Settings} & \textbf{\textsc{LLAMA3.3-70B}} \\ \midrule
Baseline & 45.58 \\
+ Solver & 63.81 \\
+ Solver \& Checker & 65.91 \\
+ Solver \& Checker \& Reflector & 71.09 \\
+ Solver \& Checker \& Reflector \& Archiver &  74.01 \\  \bottomrule
\end{tabular}%
}
\caption{Ablation study showing the incremental contribution of each agent in \multiagent.}
% \vspace{-1em}
\label{tab:ablation}
\end{table}

% Baseline & 45.58 \\
% + $\mathcal{S}$ & 63.81 \\
% + $\mathcal{S}$ \& $\mathcal{C}$ & 65.91 \\
% + $\mathcal{S}$ \& $\mathcal{C}$ \& $\mathcal{R}$ & 71.09 \\
% + $\mathcal{S}$ \& $\mathcal{C}$ \& $\mathcal{R}$ \& $\mathcal{A}$ &  \\ \bottomrule

\subsection{Memory Analysis and System Behavior}
\begin{figure}[t]
    \centering
    \includegraphics[width=0.9\columnwidth]{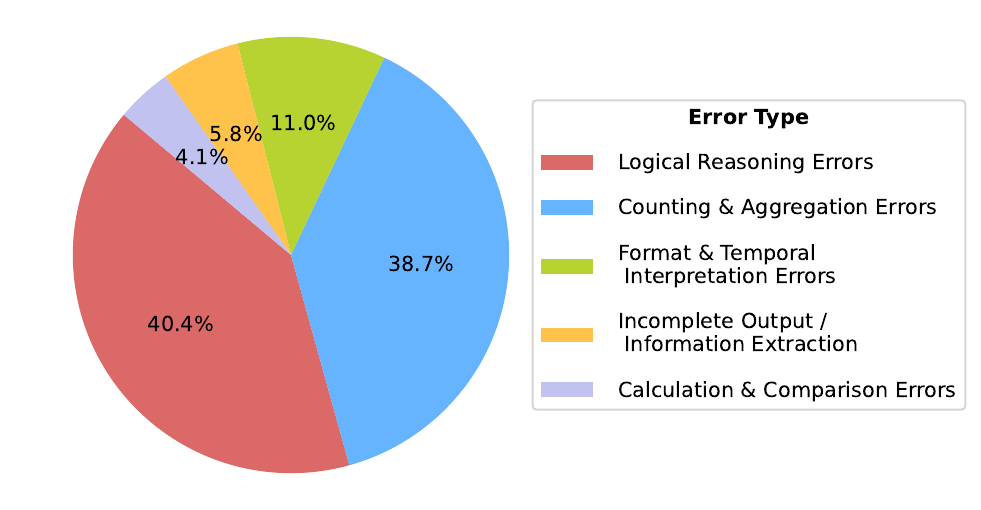}
    \vspace{-1em}
    \caption{Distribution of error types identified through \multiagent's memory system on \textsc{WikiTQ}.}
    % \vspace{-1em}
    \label{fig:error_pie}
\end{figure}

\paragraph{Error Distribution in LLM Table Reasoning.}
Figure~\ref{fig:error_pie} shows the distribution of error types on \textsc{WikiTQ}, based on all errors stored in our memory system after multiple rounds of reasoning and verification. The two most dominant categories are Logical Reasoning Errors (40.4\%) and Counting \& Aggregation Errors (38.7\%), together accounting for nearly 80\% of all failures. These are followed by Format \& Temporal Interpretation Errors (11.0\%), Incomplete Information Extraction (5.8\%), and Calculation \& Comparison Errors (4.1\%).

Notably, this distribution provides valuable insights into persistent challenges even after multi-round verification and reflection. The relatively low proportion of basic computational errors (4.1\%) suggests that our iterative verification process effectively eliminates many simpler mistakes. However, the predominance of logical reasoning and aggregation errors indicates two critical directions for future improvement: (1) enhancing the fundamental reasoning capabilities of LLMs to address the 40.4\% of logical errors, and (2) integrating specialized external tools for precise counting and aggregation operations, which could potentially resolve the 38.7\% of errors related to handling large tables with numerous entities. This analysis demonstrates how \textbf{our memory-based error categorization not only provides diagnostic information but also guides strategic research priorities for advancing table-based reasoning capabilities.}

\paragraph{Memory Dynamics and Similarity Threshold Analysis.}
\begin{table*}[t]
\centering
\resizebox{\textwidth}{!}{%
\begin{tabular}{@{}lcccccccccc@{}}
\toprule
\multicolumn{1}{l}{\textbf{Dataset}} & \textbf{\begin{tabular}[c]{@{}c@{}}Threshold\\  ($\delta$)\end{tabular}} & \textbf{\begin{tabular}[c]{@{}c@{}}Memory \\ Count\end{tabular}} & \textbf{\begin{tabular}[c]{@{}c@{}}Memory \\ Ratio (\%)\end{tabular}} & \textbf{\begin{tabular}[c]{@{}c@{}}Evolution \\ Count\end{tabular}} & \textbf{\begin{tabular}[c]{@{}c@{}}Evolution \\ Ratio (\%)\end{tabular}} & \textbf{\begin{tabular}[c]{@{}c@{}}\# Evolved \\ Memories\end{tabular}} & \textbf{\begin{tabular}[c]{@{}c@{}}Evolution \\ Efficiency ↑\end{tabular}} & \textbf{\begin{tabular}[c]{@{}c@{}}Med. Strengthen \\ Distance\end{tabular}} & \textbf{\begin{tabular}[c]{@{}c@{}}Med. Update \\ Distance\end{tabular}} & \textbf{\begin{tabular}[c]{@{}c@{}}Accuracy\\ (\%) ↑\end{tabular}} \\ \midrule
\multirow{5}{*}{\wikitq} & 0.1 & 4343 & 100.0\% & 0 & 00.0\% & 0 & 0.00 & NA & NA & 73.64 \\
 & 0.3 & 4078 & 93.9\% & 843 & 20.7\% & 981 & 1.16 & 0.25 & 0.25 & 74.01 \\
 & 0.5 & 2615 & 60.2\% & 1269 & 48.5\% & 1504 & 1.19 & 0.45 & 0.44 & 71.18 \\
 & 0.7 & 1023 & 23.5\% & 667 & 65.2\% & 820 & 1.23 & 0.64 & 0.63 &  72.28 \\
 & 0.9 & 347 & 8.0\% & 224 & 64.6\% & 254 & 1.13 & 0.81 & 0.81 & 71.82 \\
 & 1 & 191 & 4.4\% & 112 & 58.6\% & 133 & 1.19 & 0.89 & 0.88 &  71.82 \\ \midrule
\multirow{5}{*}{\textsc{TabFact}} & 0.1 & 2024 & 100.0\% & 0 & 00.0\% & 0 & 0.00 & NA & NA & 85.77 \\
 & 0.3 & 1882 & 93.0\% & 719 & 38.2\% & 787 & 1.09 & 0.24 & 0.24 & 85.70 \\
 & 0.5 & 1108 & 54.7\% & 710 & 64.1\% & 813 & 1.15 & 0.42 & 0.41 & 90.66 \\
 & 0.7 & 427 & 21.1\% & 319 & 74.7\% & 372 & 1.17 & 0.60 & 0.58 & 86.29 \\
 & 0.9 & 151 & 7.5\% & 100 & 66.2\% & 106 & 1.06 & 0.78 & 0.70 & 85.79 \\
 & 1 & 78 & 3.9\% & 53 & 67.9\% & 59 & 1.11 & 0.86 & 0.82 & 85.40 \\ \bottomrule
\end{tabular}%
}
\caption{Memory system dynamics across different similarity thresholds ($\delta$) on \wikitq~ and \textsc{TabFact} datasets.}
\vspace{-0.8em}
\label{tab:memory_statistic}
\end{table*}

To understand how similarity thresholds affect memory system behavior, we analyze memory statistics across different distance thresholds ($\delta$) on both datasets, with results presented in Table~\ref{tab:memory_statistic}.

% \vspace{-1.5mm}
\paragraph{Memory Filtering Effects.} As $\delta$ increases from 0.3 to 1.0, total memory size decreases dramatically, from 4,078 to 191 notes for \textsc{WikiTQ} and 1,882 to 78 for \textsc{TabFact}. This demonstrates our selective integration approach in action, preventing memory explosion at low thresholds while becoming increasingly selective at higher thresholds. At $\delta=0.3$, nearly all experiences (93.9\%) are added to memory, creating potential redundancy, while at $\delta=1.0$, only highly unique experiences (4.4\%) are retained, representing two extremes of the memory gradient phenomenon.

% \vspace{-1.5mm}
\paragraph{Optimal Evolution Dynamics.} The evolution ratio (percentage of memories that undergo evolution) follows an interesting pattern, peaking at approximately $\delta=0.7$ for both datasets (65.2\% for \textsc{WikiTQ}, 74.7\% for \textsc{TabFact}). Similarly, evolution efficiency, measured as evolved memories per evolution operation, reaches its maximum around the same threshold (1.23 for \textsc{WikiTQ}, 1.17 for \textsc{TabFact}). This suggests that moderate similarity thresholds create ideal conditions for knowledge evolution.

% \vspace{-1.5mm}
\paragraph{Optimal Retrieval Thresholds.} While evolution efficiency peaks at $\delta = 0.7$, optimal thresholds for accuracy show distinct patterns: \textsc{WikiTQ} achieves highest performance (74.01\%) at $\delta = 0.3$, while \textsc{TabFact} reaches peak accuracy (90.66\%) at $\delta = 0.5$. This divergence reflects the different functional requirements: during problem-solving, lower thresholds retrieve highly relevant, directly applicable memories, whereas evolution benefits from moderate thresholds balancing similarity with sufficient diversity. Performance stabilizes beyond $\delta = 0.9$, indicating a saturation point where retrieving increasingly dissimilar memories provides little additional value. These findings reveal that \textbf{$\pi_{\text{mem-retrieve}}^{\text{solver}}$ requires stricter relevance criteria (lower $\delta$ values of 0.3-0.5) for effective reasoning guidance, while $\pi_{\text{mem-retrieve}}^{\text{archiver}}$ operates optimally at $\delta = 0.7$ for efficient memory organization.}

% \vspace{-1.5mm}
\paragraph{Cross-Dataset Consistency.} Both datasets exhibit remarkably similar memory dynamics despite their different task characteristics, suggesting that these patterns reflect fundamental properties of knowledge organization rather than dataset-specific artifacts. 
% The consistency across task types provides strong evidence for the robustness of our memory evolution approach.

% \vspace{-1.5mm}
\paragraph{Theoretical Significance.} These findings align with the "approximate learning" theory in cognitive science, which posits that \textbf{optimal knowledge acquisition occurs when new information is related to existing knowledge that is neither too similar nor too different} \citep{10.1093/oxfordhb/9780195376746.013.0042}. Our empirical results showing peak evolution at moderate distances ($\delta \approx 0.7$) provide computational evidence for this cognitive principle.

This analysis reveals that memory dynamics in \multiagent~ follow a nuanced optimization pattern across different operational modes. For memory evolution, a "Goldilocks principle" applies—with too little filtering ($\delta < 0.5$), the system becomes overwhelmed with redundant information; with excessive filtering ($\delta > 0.9$), it lacks sufficient knowledge connections for meaningful evolution. The optimal range for evolution ($\delta \approx 0.7$) balances memory diversity and coherence. Meanwhile, accuracy optimization benefits from more stringent relevance criteria ($\delta = 0.3\text{-}0.5$), ensuring that only the most applicable experiences inform reasoning. This dual-threshold approach enables \multiagent~ to simultaneously optimize both knowledge organization and problem-solving performance.

\section{Conclusion}

This paper presents \multiagent, a multi-agent framework for table reasoning that integrates adaptive planning with long-term memory evolution. By decomposing reasoning into specialized functions handled by distinct agents, our approach enables dynamic strategy refinement through a feedback-driven cycle. Experiments on \wikitq~ and \textsc{TabFact} demonstrate significant improvements over existing methods, with ablation studies confirming each component's value. Our memory analysis reveals that logical reasoning errors and counting/aggregation operations account for nearly 80\% of remaining mistakes, suggesting 2 promising directions: enhancing fundamental reasoning capabilities and developing specialized numerical tools for complex operations. Beyond table reasoning, the principles demonstrated in \multiagent~ may benefit knowledge-intensive tasks where verification, reflection, and experience accumulation are crucial.
\section*{Limitations}
Despite \multiagent's promising results, several limitations should be acknowledged. First, our approach is computationally more intensive than single-pass methods due to its multi-round, multi-agent architecture. Each reasoning attempt requires multiple LLM calls across different agents, increasing both inference time and computational costs. This presents challenges for real-time applications or deployment on resource-constrained systems.

% Second, \multiagent's performance gains come with increased prompt complexity. The specialized prompts for each agent require careful engineering, potentially limiting the framework's transferability to new domains without significant adaptation. This dependency on prompt design could affect the system's robustness when confronted with questions or tables that differ substantially from those in our evaluation datasets.

Second, while our memory evolution mechanism demonstrates effectiveness in our experiments, its long-term scalability remains unexplored. As the memory base grows, maintaining coherence and preventing knowledge dilution become increasingly challenging. Future work should examine more sophisticated memory management strategies, including forgetting mechanisms and hierarchical organization of memory notes.

Finally, our framework currently focuses exclusively on table-based reasoning without incorporating external knowledge. This limits its applicability to questions requiring information beyond what's explicitly presented in the table. Enhancing \multiagent with external knowledge collection capabilities would be a valuable extension to address this limitation.

% \section*{Acknowledgments}

\bibliography{custom}

\clearpage
\section*{Appendix}
\label{sec:appendix}
\appendix

\section{Related Work}
\label{related_work_appendix}
\begin{table*}[t]
\centering
\resizebox{\textwidth}{!}{%
\begin{tabular}{@{}lccccccc@{}}
\toprule
\textbf{Model} & \textbf{\begin{tabular}[c]{@{}c@{}}Multi-Agent\\ System\end{tabular}} & \textbf{\begin{tabular}[c]{@{}c@{}}External\\ Tool Use\end{tabular}} & \textbf{ReAct} & \textbf{Reflection} & \textbf{\begin{tabular}[c]{@{}c@{}}Self-\\ Refinement\end{tabular}} & \textbf{\begin{tabular}[c]{@{}c@{}}Long-Term\\ Memory\end{tabular}} & \textbf{\begin{tabular}[c]{@{}c@{}}Dynamic\\ Planning\end{tabular}} \\ \midrule
Binder~\citep{cheng_binding_2023} & \textcolor{red}{\ding{55}} & \textcolor{green}{\ding{51}} & \textcolor{red}{\ding{55}} & \textcolor{red}{\ding{55}} & \textcolor{red}{\ding{55}} & \textcolor{red}{\ding{55}} & \textcolor{red}{\ding{55}} \\
Dater~\citep{ye_large_2023} & \textcolor{green}{\ding{51}} & \textcolor{green}{\ding{51}} & \textcolor{red}{\ding{55}} & \textcolor{red}{\ding{55}} & \textcolor{red}{\ding{55}} & \textcolor{red}{\ding{55}} & \textcolor{red}{\ding{55}} \\
Chain-of-Table~\citep{wang_chain--table_2024} & \textcolor{red}{\ding{55}} & \textcolor{red}{\ding{55}} & \textcolor{green}{\ding{51}} & \textcolor{red}{\ding{55}} & \textcolor{red}{\ding{55}} & \textcolor{red}{\ding{55}} & \textcolor{red}{\ding{55}} \\
ReAcTable~\citep{zhang_reactable_2023} & \textcolor{red}{\ding{55}} & \textcolor{green}{\ding{51}} & \textcolor{green}{\ding{51}} & \textcolor{red}{\ding{55}} & \textcolor{red}{\ding{55}} & \textcolor{red}{\ding{55}} & \textcolor{red}{\ding{55}} \\
Table-Critic~\citep{yu2025tablecriticmultiagentframeworkcollaborative} & \textcolor{green}{\ding{51}} & \textcolor{red}{\ding{55}} & \textcolor{red}{\ding{55}} & \textcolor{green}{\ding{51}} & \textcolor{green}{\ding{51}} & \textcolor{green}{\ding{51}} & \textcolor{red}{\ding{55}} \\ \midrule
\textbf{MAPLE (Ours)} & \textcolor{green}{\ding{51}} & \textcolor{red}{\ding{55}} & \textcolor{green}{\ding{51}} & \textcolor{green}{\ding{51}} & \textcolor{green}{\ding{51}} & \textcolor{green}{\ding{51}} & \textcolor{green}{\ding{51}} \\ \bottomrule
\end{tabular}%
}
\caption{Comparison of representative table reasoning methods across key design dimensions.}
\label{tab:method_compare}
\end{table*}

As illustrated in Table~\ref{tab:method_compare}, existing table reasoning methods exhibit significant limitations across key design dimensions. While some implement multi-agent architectures or ReAct-based reasoning, none integrates all critical components: dynamic planning, reflection mechanisms, self-refinement, and long-term memory. Our proposed \multiagent~ framework addresses these gaps by combining collaborative verification, adaptive planning, and evolving memory structures to achieve more robust and accurate table reasoning.

% {\small
\begin{algorithm*}[t]
\caption{:~\multiagent's Dynamic Memory Evolution Process}
\label{alg:memory_pipeline}
% \begingroup
% \setstretch{0.85}
\begin{algorithmic}[1]
\State \textbf{Input:} Working Memory $\mathcal{M}_w = \{\mathcal{T}, q, a_m, a_g, \tau, d, p\}$, Long-term Memory $\mathcal{M}_l$, Distance Threshold $\delta$, Neighbor Limit $k$, Required Minimum Neighbors $k_{min}$
\State \textbf{Output:} Updated Long-term Memory $\mathcal{M}_l$

\For{each sample in working memory}
    \State $m \gets \pi_{\text{archiver-sum}}(\mathcal{M}_w)$ \Comment{Distill experience into memory note, Eq.~\ref{eq:advisor_sum}}
    \State $\mathcal{N}_{\text{archiver}} \gets$ $\pi_{\text{mem-retrieve}}^{\text{archiver}}(m , \mathcal{M}_l, k, \delta)$ \Comment{Retrieve $\leq k$ similar memories within $\delta$, Eq.~\ref{eq:advisor_re_advisor}}
    \If{$|\mathcal{N}_{\text{archiver}}| < k_{min}$} \Comment{Filter redundant memories}
        \State $e \gets \pi_{\text{archiver-evo}}(m, \mathcal{N}_{\text{archiver}})$ \Comment{Decide evolution actions, Eq.~\ref{eq:advisor_evol}} 
        \If{$e.\texttt{should\_evolve}$ == True}
            \If{$\texttt{strengthen} \in e.\texttt{actions}$}
                \State \textsc{AddLinks}$(m, e.\texttt{suggested\_connections})$ \Comment{Create semantic connections}
            \EndIf
            \If{$\texttt{update\_neighbor} \in e.\texttt{actions}$}
                \State \textsc{UpdateMetadata}$(\mathcal{N}_{\text{archiver}}, e.\texttt{new\_info\_neighbor}$ \Comment{Refine neighbor metadata}
            \EndIf
            \State $m.\texttt{tags} \gets e.\texttt{tags\_to\_update}$ \Comment{Update semantic tags}
        \EndIf
        \State \textsc{AddMemory}$(\mathcal{M}_l, m)$ \Comment{Persist new experience}
    \EndIf
\EndFor
\State \Return $\mathcal{M}_l$
\end{algorithmic}
% \endgroup
\end{algorithm*}
% }

\section{Cognitive Architecture}
\begin{figure*}[t]
    \centering
    \includegraphics[width=\textwidth]{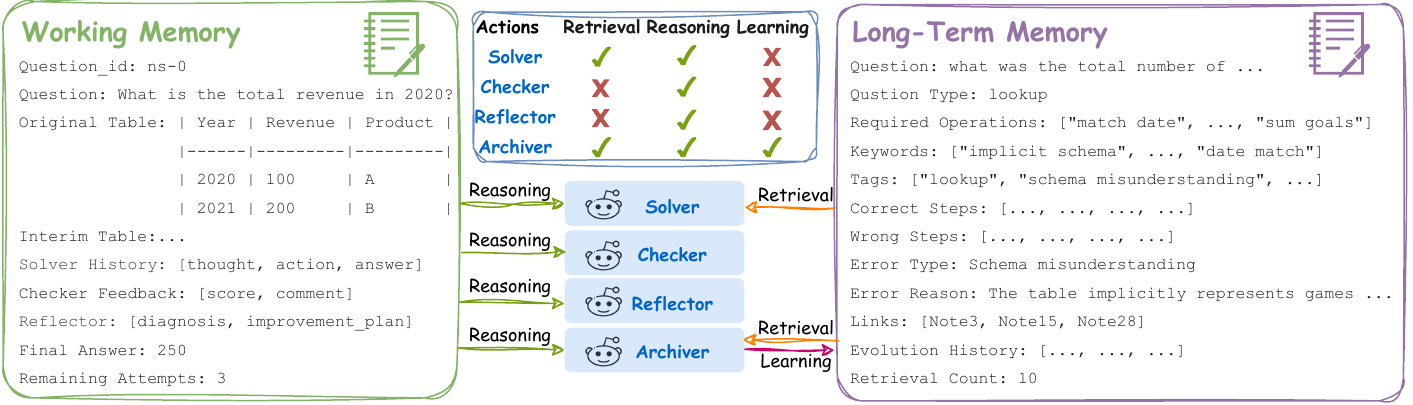}
    \caption{Overview of the memory structures and information flows in \multiagent. 
    % The working memory stores transient task-specific information, including the original table, question, Solver action history, Checker feedback, and Reflector suggestions. The long-term memory maintains structured historical knowledge such as reasoning patterns, error types, and solution strategies. 
    The \textcolor{teal}{green arrows ($\rightarrow$)} represent reasoning processes, where agents \textbf{read and update} working memory during multi-step problem solving. The \textcolor{orange}{orange arrows ($\leftarrow$)} represent \textbf{retrieval} operations from long-term memory to support current reasoning. The \textcolor{purple}{red arrows ($\rightarrow$)} denote learning operations, where new knowledge is \textbf{written back} into the long-term memory.
    % enabling experience accumulation across tasks. This bidirectional flow creates a continuous cycle of knowledge utilization and evolution that distinguishes MAPLE from traditional static reasoning systems.
    }
    \label{fig:ModelAction}
\end{figure*}

\subsection{Memory Module}
\label{appendix_memory}
To enable multi-step reasoning, verification, and reflection, \multiagent~ organizes internal information across two complementary memory modules: a short-term working memory and a long-term memory. The working memory enables flexible planning and adaptation during reasoning by dynamically maintaining intermediate states, while the long-term memory provides stable knowledge accumulated across tasks to guide future decisions. Together, these memory structures allow different agents to persist, access, and manipulate relevant information throughout and across problem-solving sessions.

\paragraph{Working Memory.}

The working memory $(\mathcal{M}_w)$ temporarily stores all information related to the current task instance, implementing a Shared Message Pool architecture~\citep{hong2024metagptmetaprogrammingmultiagent} for agent communication. As shown in Figure~\ref{fig:ModelAction}, it maintains the original table $\mathcal{T}$, the question $q$, the Solver's operation history (including intermediate tables and tentative answers), Checker feedback (scores and comments), Reflector analysis (diagnosis and suggestions), and task-level metadata such as the number of remaining attempts.

Unlike centralized or hierarchical communication structures, our Shared Message Pool enables all agents to asynchronously publish information to and subscribe from a common memory space. This architecture facilitates flexible many-to-many interactions without predefined communication pathways, allowing emergent collaboration patterns based on informational dependencies rather than rigid control flow. For example, the Reflector can simultaneously observe both Solver reasoning steps and Checker feedback, synthesizing insights that would be difficult to achieve in strictly layered or peer-to-peer architectures.

To directly facilitate communication with large language models during multi-turn interactions, the working memory is represented entirely in natural language format. Each agent, Solver, Checker, Reflector, and Archiver, reads from and writes to this shared memory throughout the reasoning cycle, ensuring that context is consistently updated and accessible at every decision point.

\paragraph{Long-term Memory.}

The long-term memory $(\mathcal{M}_l)$ captures accumulated knowledge across tasks, supporting continual improvement and experience reuse. Inspired by frameworks like A-MEM~\citep{xu2025amemagenticmemoryllm}, we adapt their approaches specifically for table-based reasoning challenges.

Our memory structure is tailored for table QA, with each memory note containing fields crucial for reasoning diagnostics: question type, required operations, correct/wrong steps, error types, reasons and additional metadata enables tracking of memory dynamics over time. We implement a hybrid storage format—structured metadata for embedding-based retrieval and natural language descriptions for interpretability.

A key innovation is our selective integration mechanism, which differs from frameworks like A-MEM that accumulate all experiences indiscriminately. While A-MEM addresses long-range conversations where preserving every detail is essential, table QA questions often exhibit high similarity and structural homogeneity, making comprehensive retention inefficient. Our approach implements density-based filtering: when a new memory note closely resembles multiple existing entries, it is not added to the repository, preventing memory saturation while preserving diverse reasoning patterns.

Similarly, while adopting the concept of memory evolution from prior work, we introduce strategic timing for evolution operations, performing them selectively rather than universally, to maximize coherence without unnecessary computational overhead.

This specialized approach achieves an optimal balance between knowledge retention and efficiency for table reasoning tasks. During operation, the Archiver retrieves relevant memories to assist the Solver, and after task completion, strategically updates the memory base by either inserting unique experiences or evolving existing entries to strengthen their utility.

\subsection{Action Module}
The action module is responsible for translating an agent's decisions into specific outcomes. Positioned at the most downstream point of the architecture, it directly interacts with the environment and is influenced by the memory and planning modules.

Following the categorization proposed in CoALA Framework~\citep{sumers2024cognitivearchitectureslanguageagents}, action spaces can be broadly divided into two categories: \textit{external actions} and \textit{internal actions}. External actions involve interactions with the external environment, such as controlling a robot, communicating with a human, or navigating a website. As our framework focuses on table reasoning tasks, we do not involve external actions.

Instead, our method operates entirely within internal action spaces, where actions are directed toward interacting with internal memory systems. Internal actions can be further classified into three types based on their interaction with memory: \textbf{Retrieval} involves reading from long-term memory to access relevant past experiences. \textbf{Reasoning} refers to updating the short-term working memory through LLM-based processing of current information. \textbf{Learning} denotes writing new information into long-term memory for future use.

These fundamental actions rarely occur in isolation, instead, they form characteristic sequences and combinations that enable sophisticated reasoning patterns. For example, the Solver typically engages in iterative reasoning cycles punctuated by occasional retrieval operations, while the Archiver combines retrieval and learning to maintain memory coherence. The power of our multi-agent approach emerges from these diverse action patterns, allowing different agents to specialize in distinct cognitive operations while maintaining a cohesive problem-solving process.

\Cref{fig:ModelAction} summarizes the internal actions associated with each agent in our system. Both the Solver and Checker agents primarily engage in reasoning actions: they read from the working memory, process information according to their designated roles, and write the updated reasoning steps back into the working memory. The Reflector agent performs both reasoning and learning, as it not only updates the working memory but also contributes insights to the long-term memory. The Archiver agent engages in all three action types: retrieval to access relevant experiences, reasoning to analyze current tasks, and learning to evolve the memory base with new knowledge.

\subsection{Planning Module}

Effective planning is crucial for solving multi-step reasoning tasks, where the sequence and selection of actions can significantly impact final outcomes. Following the categorization proposed by~\citet{Wang_2024}, planning approaches can be broadly classified into two categories: planning without feedback and planning with feedback.

\paragraph{Planning without Feedback.} Traditional reasoning systems typically employ static planning, where the entire reasoning trajectory is predetermined before execution. For instance, standard Chain-of-Thought prompting generates all reasoning steps in a single forward pass without adjusting to intermediate results. Whether using single-path reasoning (where each step leads to exactly one subsequent step) or multi-path reasoning (where reasoning steps form tree-like structures), these approaches struggle with complex tasks where initial plans require revision based on unexpected discoveries during execution. The fundamental limitation is their inability to iteratively refine strategies based on execution outcomes—a critical capability in human problem-solving.

\paragraph{Planning with Feedback.} In contrast, \multiagent~ implements dynamic planning with dual-source feedback, enabling adaptive reasoning that more closely mirrors human cognition:

\textbf{Environmental Feedback} enables the Solver to observe changes in the table state after each operation and decide whether to continue manipulation or derive an answer. Similar to approaches like ReAct~\citep{yao2023reactsynergizingreasoningacting}, our framework incorporates thought-action-observation triplets, allowing the Solver to adapt its reasoning trajectory based on real-time observations of how table manipulations affect the environment state. This environmental grounding prevents the accumulation of reasoning errors that plague single-pass methods.

\textbf{Model Feedback} from specialized verification agents (Checker and Reflector) provides structured evaluation of reasoning quality. Unlike self-reflection approaches where the same model instance both generates and evaluates its own solutions, our architecture implements a clear separation of concerns, dedicated agents with specialized prompts and evaluation criteria perform verification tasks. This functional modularity enables more objective assessment, as the Checker evaluates answers without access to the generation process, and the Reflector provides targeted diagnosis rather than mere self-justification. This division of cognitive labor creates a system of checks and balances that significantly reduces the confirmation bias inherent in single-model reflection approaches.

As illustrated in Figure~\ref{fig:ModelFramework}, this feedback-driven planning eliminates the need for predefined reasoning sequences. Instead, the exact path through the reasoning space emerges dynamically from agent interactions: the Solver adjusts based on intermediate table states and Reflector diagnostics, the Checker determines when reasoning quality meets acceptance criteria, and the Archiver retrieves relevant experiences to guide initial approaches. This distributed, adaptive planning architecture creates an output-feedback-refinement loop that iteratively improves reasoning quality—a capability fundamental to robust problem-solving but absent in traditional single-pass systems.

\section{Memory Evolution Algorithm}
\label{app:memory_algorithm}

In this section, we present the detailed algorithm for \multiagent's memory evolution process (\Cref{alg:memory_pipeline}). While the main text describes the conceptual framework and key innovations of our memory system, this appendix provides the complete algorithmic implementation of how new experiences are evaluated, filtered, and integrated into the long-term memory base. Specifically, the \textsc{AddLinks} function corresponds to the \emph{Strengthen} option in \Cref{fig:Advisor_evol}: it does not alter the metadata of the neighboring memories but instead appends their \texttt{memory\_id} to the \texttt{Links} field of the current memory. In contrast, the \textsc{UpdateMetadata} function corresponds to the \emph{Update\_Neighbor} option in \Cref{fig:Advisor_evol}: it summarizes new \texttt{tags} and \texttt{context} based on both the current and neighboring memories, and updates these fields in each memory accordingly.

\section{Case Study}
\label{appendix_case_study}
\begin{figure}[t]
\centering
\begin{tabbing}
\hspace{4em} \= \hspace{10em} \= \kill
Question \& Table  \\
\> $\downarrow$ \\
Solver Round 1~(\Cref{round1_s})  \\
\> $\downarrow$ \\
Answer: Clint Dempsey \\
\> $\downarrow$ \\
Checker Round 1~(\Cref{round2_c}) \\  
\> $\downarrow$ \\
Logic Error \textcolor{red}{\ding{55}} \\
\> $\downarrow$ \\
Reflector~(\Cref{round3_r}) \\
\> $\downarrow$ \\
Solver Round 2~(\Cref{round4_s},~\ref{round5_s}) \\
\> $\downarrow$ \\
Final Answer: Eric Wynalda \\
\> $\downarrow$ \\
Checker Round 2~(\Cref{round6_c})  \\
\> $\downarrow$ \\
The answer is correct \textcolor{green}{\ding{51}} \\
\> $\downarrow$ \\
Archiver~(\Cref{round7_a},~\ref{round8_a})  \\
\> $\downarrow$ \\
Save to Memory Base
\end{tabbing}
\caption{Illustrative case study of MAPLE's multi-agent reasoning workflow.}
\label{fig:case_study_flow}
\end{figure}

To illustrate how \multiagent’s agents collaborate to refine reasoning, we present a step-by-step case study. As shown in Figure~\ref{fig:case_study_flow}, the Solver begins with an initial attempt based on the input table and question. After generating an intermediate answer, the Checker detects a logical error, prompting the Reflector to diagnose the mistake and provide actionable feedback. Incorporating these insights, the Solver re-attempts the task and successfully derives the correct answer in the next round. The final reasoning trace is then passed to the Archiver, which summarizes it into a memory note and evaluates whether to evolve the memory base. In this case, the Archiver decides that no evolution is necessary.

\subsection{Round 1 - Solver}
\label{round1_s}
\subsubsection{User Input}
Below is a retrieved memory from a similar question previously answered by the system. You may use it as a reference to inform your reasoning.

<Related Memory>
\begin{itemize}[itemsep=0.001em, topsep=0pt, leftmargin=1.5em]
  \item Past Question: who scored the most goals?
  \item Question Type: aggregation
  \item Required Operations: find maximum, compare
  \item Correct Reasoning Steps: 1.Identify the column representing the number of goals scored by each player; 2.Find the maximum value in the 'Goals' column; 3.Return the player associated with the maximum number of goals;
  \item Error Type: none
  \item Error Reason: none
\end{itemize}

Now, here is your actual Solver task. This is your 1 attempt. You have 4 attempts remaining.

<Table>

See~\Cref{tab:example_table}
\begin{table}[t]
\centering
\resizebox{\columnwidth}{!}{%
\begin{tabular}{@{}clrrl@{}}
\toprule
\textbf{\#} & \textbf{Player} & \textbf{Goals} & \textbf{Caps} & \textbf{Career} \\ \midrule
1 & Landon Donovan & 57 & 155 & 2000--present \\
2 & Clint Dempsey & 36 & 103 & 2004--present \\
3 & Eric Wynalda & 34 & 106 & 1990--2000 \\
4 & Brian McBride & 30 & 95 & 1993--2006 \\
5 & Joe-Max Moore & 24 & 100 & 1992--2002 \\
6T & Jozy Altidore & 21 & 67 & 2007--present \\
6T & Bruce Murray & 21 & 86 & 1985--1993 \\
8 & Eddie Johnson & 19 & 62 & 2004--present \\
9T & Earnie Stewart & 17 & 101 & 1990--2004 \\
9T & DaMarcus Beasley & 17 & 114 & 2001--present \\ \bottomrule
\end{tabular}%
}
\caption{Original input table used for case study. This is the table provided to the Solver at the beginning of the reasoning process.}
\label{tab:example_table}
\end{table}

<Question>

who was the top goalscorer previous to landon donovan?

\subsubsection{LLM Response}
\begin{itemize}[itemsep=0.001em, topsep=0pt, leftmargin=1.5em]
  \item Thought: To find the top goalscorer before Landon Donovan, I need to identify the player with the most goals scored, excluding Landon Donovan. Since the table is already sorted by the number of goals in descending order, I can look at the second row to find the player with the next highest number of goals.
  \item Action: Identify the player with the second-highest number of goals
  \item Intermediate table: <NOT CHANGED>
  \item Answer: Clint Dempsey
\end{itemize}

\subsection{Round 2 - Checker}
\label{round2_c}
\subsubsection{User Input}
<Table>

Original Table (\Cref{tab:example_table})

<Question>

who was the top goalscorer previous to landon donovan?

<Answer>

Clint Dempsey

\subsubsection{LLM Response}
\paragraph{Answer Type Checking}
\begin{itemize}[itemsep=0.001em, topsep=0pt, leftmargin=1.5em]
  \item Score: 2
  \item Comments: The question asks for a player's name, and the answer is a player's name, so the type matches correctly.
\end{itemize}

\paragraph{Format Validation}
\begin{itemize}[itemsep=0.001em, topsep=0pt, leftmargin=1.5em]
  \item Score: 2
  \item Comments: The answer is a single player's name, which follows the expected format for this type of question.
\end{itemize}

\paragraph{Evidence Grounding}
\begin{itemize}[itemsep=0.001em, topsep=0pt, leftmargin=1.5em]
  \item Score: 0
  \item Comments: According to the table, before Landon Donovan, the top goalscorer would be the player with the next highest number of goals. Clint Dempsey has 36 goals, but Eric Wynalda has 34 goals and is listed above Dempsey in the table in terms of goals. However, the correct answer should be the one with the highest goals before Donovan, which is not Dempsey but actually the player with the next highest goal count after Donovan is not the correct interpretation. The correct interpretation is to find the player who was the top scorer before Donovan's time or the one with the next highest goals if considering the table's data. Since Donovan has 57 goals, the next in line is Dempsey with 36 goals, but considering the question's phrasing, it seems to ask for the top scorer before Donovan's emergence or record, which would indeed be Eric Wynalda or another player based on the 'previous to' criteria. Given the table, Eric Wynalda had 34 goals and was a top scorer before Donovan's time, making Dempsey not the correct answer based on the 'previous to' interpretation.
\end{itemize}

\paragraph{Summary}
\begin{itemize}[itemsep=0.001em, topsep=0pt, leftmargin=1.5em]
  \item Total Score: 4
  \item Final Comments: The answer is incorrect because, based on the logical interpretation of 'previous to Landon Donovan', the correct top goalscorer before Donovan would likely be Eric Wynalda, given the context and data provided in the table. Clint Dempsey does have a significant number of goals but does not fit the 'previous to' criteria as directly as another player might.
\end{itemize}

\subsection{Round 3 - Reflector}
\label{round3_r}
\subsubsection{User Input}
<Question> 

who was the top goalscorer previous to landon donovan?

<Table>

Original Table (\Cref{tab:example_table})

<Reasoner’s Processing History>

1.Identify the player with the second-highest number of goals

<Reasoner’s Final answer>

Clint Dempsey

<Checker Feedback>

Same as Round 2~(\ref{round2_c}) Checker's output 

\subsubsection{LLM Response}
\begin{itemize}[itemsep=0.001em, topsep=0pt, leftmargin=1.5em]
  \item Diagnosis: The reasoner incorrectly identified Clint Dempsey as the top goalscorer before Landon Donovan. The mistake lies in the interpretation of 'previous to', which implies considering the time or emergence of players, not just the next highest goal count. The reasoner should have considered the career span and goals of players who were active before or alongside Donovan's early career to accurately determine the top scorer before him.
  \item Improvement plan: To improve, the reasoner should re-evaluate the question's phrasing and the table's data. Specifically, it should identify players who were active before Landon Donovan's emergence or during his early career and then determine which of these players had the highest number of goals. In this case, considering Eric Wynalda's career span (1990-2000) and goal count (34), he would be a more appropriate answer as the top goalscorer before Donovan, given the 'previous to' criteria. The reasoner needs to enhance its ability to interpret temporal relationships and contextual clues within the question to provide a more accurate answer.
\end{itemize}

\subsection{Round 4 - Solver}
\label{round4_s}
\subsubsection{User Input}
Below is a retrieved memory from a similar question previously answered by the system. You may use it as a reference to inform your reasoning.

<Related Memory>

Same memory content as in Round 1(\ref{round1_s}) Solver input.

Now, here is your actual Solver task. This is your 1 attempt. You have 4 attempts remaining. (Note: This refers to the inner-round count under the ReAct paradigm, not the outer round of Solver retries.)

<Table>

Original Table (\Cref{tab:example_table})

<Question>

who was the top goalscorer previous to landon donovan?

If the <Reflector result> is available, it means you previously made a mistake, and the Reflector has summarized the cause of that error. So this time, when solving the question, please take the improvement plan section into account to avoid making the same mistake again.

<Reflector Result>

Same as Round 3(\ref{round3_r}) Reflector's output

\begin{table}[t]
\centering
\resizebox{\columnwidth}{!}{%
\begin{tabular}{@{}clrrl@{}}
\toprule
\textbf{\#} & \textbf{Player} & \textbf{Goals} & \textbf{Caps} & \textbf{Career} \\ \midrule
3 & Eric Wynalda & 34 & 106 & 1990--2000 \\
4 & Brian McBride & 30 & 95 & 1993--2006 \\
5 & Joe-Max Moore & 24 & 100 & 1992--2002 \\
6T & Bruce Murray & 21 & 86 & 1985--1993 \\
9T & Earnie Stewart & 17 & 101 & 1990--2004 \\ \bottomrule
\end{tabular}%
}
\caption{Intermediate table produced by the Solver during the reasoning process. Irrelevant rows have been filtered out based on the question.}
\label{tab:example_table_2}
\end{table}
\subsubsection{LLM Response}
\begin{itemize}[itemsep=0.001em, topsep=0pt, leftmargin=1.5em]
  \item Thought: The question asks for the top goalscorer previous to Landon Donovan. The Reflector result indicates that I should consider the career span and goals of players who were active before or alongside Donovan's early career. Landon Donovan's career started in 2000. I need to identify players who were active before 2000 and find the one with the highest number of goals.
  \item Action: Filter players whose career ended before 2000 or started before 2000, then find the maximum goals among them
  \item Intermediate table: see~\Cref{tab:example_table_2}
  \item Answer: <NOT READY>
\end{itemize}

\subsection{Round 5 - Solver}
\label{round5_s}
\subsubsection{User Input}
Below is a retrieved memory from a similar question previously answered by the system. You may use it as a reference to inform your reasoning.

<Related Memory>

Same memory content as in Round 1(\ref{round1_s}) Solver input.

Now, here is your actual Solver task. This is your 2 attempt. You have 3 attempts remaining. 

<Table>

Intermediate Table(\Cref{tab:example_table_2}) from Round 4(\ref{round4_s}) 

<Question>

who was the top goalscorer previous to landon donovan?

<Action History>

1.Filter players whose career ended before 2000 or started before 2000, then find the maximum goals among them

<Reflector Result>

Same as Round 3(\ref{round3_r}) Reflector's output

\subsubsection{LLM Response}
\begin{itemize}[itemsep=0.001em, topsep=0pt, leftmargin=1.5em]
  \item Thought: The Reflector result indicates that I should consider players who were active before or during Landon Donovan's early career. Given the 'previous to' criteria, I need to identify the top goalscorer among players whose career span ended before or overlapped with Donovan's emergence. Eric Wynalda's career (1990-2000) and high goal count (34) make him a strong candidate. I will re-evaluate the table data to confirm this.
  \item Action: Identify top goalscorer among players active before or during Landon Donovan's early career
  \item Intermediate table: <NOT CHANGED>
  \item Answer: Eric Wynalda
\end{itemize}

\subsection{Round 6 - Checker}
\label{round6_c}
\subsubsection{User Input}
<Table>

Original Table (\Cref{tab:example_table})

<Question>

who was the top goalscorer previous to landon donovan?

<answer>

Eric Wynalda

\subsubsection{LLM Response}
\paragraph{Answer Type Checking}
\begin{itemize}[itemsep=0.001em, topsep=0pt, leftmargin=1.5em]
  \item Score: 2
  \item Comments: The question asks for a player's name, and the answer is a name, which matches the expected type.
\end{itemize}

\paragraph{Format Validation}
\begin{itemize}[itemsep=0.001em, topsep=0pt, leftmargin=1.5em]
  \item Score: 2
  \item Comments: The answer is a single player's name, which follows the expected format for this type of question.
\end{itemize}

\paragraph{Evidence Grounding}
\begin{itemize}[itemsep=0.001em, topsep=0pt, leftmargin=1.5em]
  \item Score: 2
  \item Comments: According to the table, before Landon Donovan, Eric Wynalda was indeed the top goalscorer with 34 goals, which is less than Donovan's 57 but more than the others below him in the list.
\end{itemize}

\paragraph{Summary}
\begin{itemize}[itemsep=0.001em, topsep=0pt, leftmargin=1.5em]
  \item Total Score: 6
  \item Final Comments: The answer is correct in terms of type, format, and evidence grounding. Eric Wynalda is the player with the most goals before Landon Donovan, making the response accurate.
\end{itemize}

\subsection{Round 7 - Archiver (Memory Summarization)}
\label{round7_a}
\subsubsection{User Input}
<Question>

who was the top goalscorer previous to landon donovan?

<Table>

Original Table (\Cref{tab:example_table})

<Model Answer>

Eric Wynalda

<Ground Truth>

Eric Wynalda

<Reasoner History>

1. Filter players whose career ended before 2000 or started before 2000, then find the maximum goals among them

2. Identify top goalscorer among players active before or during Landon Donovan's early career

<Reflector Feedback>

Same memory content as in Round 3(\ref{round3_r}) Reflector's output.

\subsubsection{LLM Response}
\begin{itemize}[itemsep=0.001em, topsep=0pt, leftmargin=1.5em]
  \item Question Type: lookup
  \item Required Operations: ['filter', 'compare', 'identify max']
  \item Context: This question requires identifying the top goalscorer before Landon Donovan's time, involving filtering players based on their career timeline and comparing their goal scores. The correct answer, Eric Wynalda, is determined by having the highest number of goals among players whose careers significantly predated or overlapped with Donovan's start in 2000.
  \item Keywords: ['filter by time', 'compare scores', 'max goals']
  \item Tags: ['lookup', 'sports data', 'goalscorer', 'career timeline']
  \item Correct Steps: ["Identify players with careers before or overlapping Landon Donovan's start in 2000", 'Among these players, find the one with the highest number of goals', 'Return the name of this top goalscorer']
  \item Wrong Steps: [ ]
  \item Error Type: none
  \item Error Reason: none
\end{itemize}

\subsection{Round 8 - Archiver (Memory Evolution)}
\label{round8_a}
\subsubsection{User Input}
<New Memory Content>
\begin{itemize}[itemsep=0.001em, topsep=0pt, leftmargin=1.5em]
  \item Question ID: nu-2024
  \item Question: who was the top goalscorer previous to landon donovan?
  \item Question Type: lookup
  \item Required operations: filter, compare, identify max
  \item Context: This question requires identifying the top goalscorer before Landon Donovan's time, involving filtering players based on their career timeline and comparing their goal scores. The correct answer, Eric Wynalda, is determined by having the highest number of goals among players whose careers significantly predated or overlapped with Donovan's start in 2000.
  \item Tags: lookup, sports data, goalscorer, career timeline
  \item Keywords: filter by time, compare scores, max goals
  \item Correct Answer: Eric Wynalda
  \item Model Answer: Eric Wynalda
  \item Correct Steps: - Identify players with careers before or overlapping Landon Donovan's start in 2000; - Among these players, find the one with the highest number of goals; - Return the name of this top goalscorer
  \item Error Type: none
  \item Error Reason: none
\end{itemize}

<Nearest Neighbors Memories>

Memory Note 1; Memory Note 2. (The format and content are similar to the memory note shown above.)

\subsubsection{LLM Response}
\begin{itemize}[itemsep=0.001em, topsep=0pt, leftmargin=1.5em]
  \item Should Evolve: false
  \item Actions: [ ]
  \item Suggested Connections: [ ]
  \item Tags to Update: [ ]
  \item New Context Neighborhood: [ ]
  \item New Tags Neighborhood: [ ]
\end{itemize}

\section{Addtional Experimental Results}
\label{appendix_additional_exp}
\subsection{Table Size vs. Reasoning Performance}
\label{table_size_appendix}
\begin{figure}[t]
    \centering
    \includegraphics[width=\columnwidth]{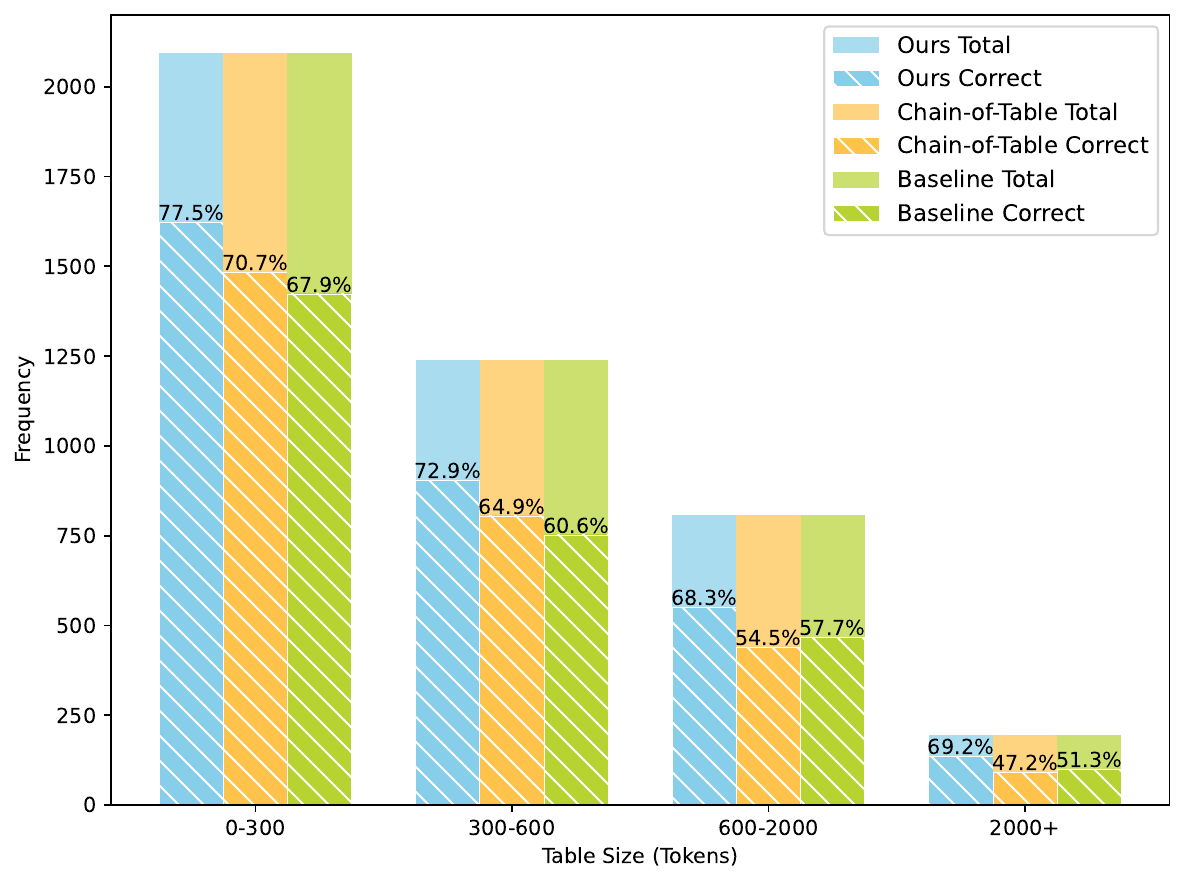}
    \caption{Accuracy comparison across table size categories on \textsc{WikiTQ}. Performance is shown for \multiagent~ (\textcolor{CornflowerBlue}{blue}), Chain-of-Table (\textcolor{orange}{orange}), and Chain-of-Thought baseline (\textcolor{ForestGreen}{green}), with both total attempt counts (darker shade) and correct answers (lighter stripe pattern) displayed for each method. }
    \label{fig:table_size}
\end{figure}
We analyze how table size affects reasoning performance on the \textsc{WikiTQ} dataset by categorizing tables into four buckets based on token length: 0–300, 300–600, 600–2000, and 2000+. Figure~\ref{fig:table_size} presents the accuracy trends for \multiagent~ (\textcolor{CornflowerBlue}{blue}), Chain-of-Table (\textcolor{orange}{orange}), and a Chain-of-Thought baseline (\textcolor{ForestGreen}{green}) across these size categories.

Two key observations emerge. First, as table size increases, all methods experience a performance decline. This aligns with the intuition that larger tables introduce greater information complexity and noise, making it harder for LLMs to extract relevant content effectively. Second, \multiagent~ consistently outperforms both baselines across all size ranges, with particularly strong gains on larger tables (600+ tokens). For instance, in the 600–2000 range, \multiagent~ achieves 68.3\% accuracy, compared to 54.5\% and 57.7\% for Chain-of-Table and the CoT baseline, representing relative improvements of 13.8\% and 10.6\%. Notably, while the performance gap between methods narrows for the smallest tables, it widens substantially as complexity increases, suggesting that \multiagent's adaptive multi-agent architecture and memory-guided planning provide crucial robustness against information overload in complex tabular contexts.

% \begin{figure}[htbp]
%   \centering
%   \begin{subfigure}[b]{0.49\columnwidth}
%     \includegraphics[width=\linewidth]{figures/wiki_iteration_stats.pdf}
%     \caption{\textsc{WikiTQ}}
%     \label{fig:wiki}
%   \end{subfigure}
%   \hfill
%   \begin{subfigure}[b]{0.49\columnwidth}
%     \includegraphics[width=\linewidth]{figures/tab_iteration_stats.pdf}
%     \caption{\textsc{TabFact}}
%     \label{fig:tabfact}
%   \end{subfigure}
%   \caption{Iteration vs. accuracy curves on two datasets. The bar plots represent the frequency of iterations, while the orange line shows cumulative accuracy at each iteration count.}
%   \label{fig:iter_acc}
% \end{figure}

\begin{figure}[t]
  \centering
  \includegraphics[width=\columnwidth]{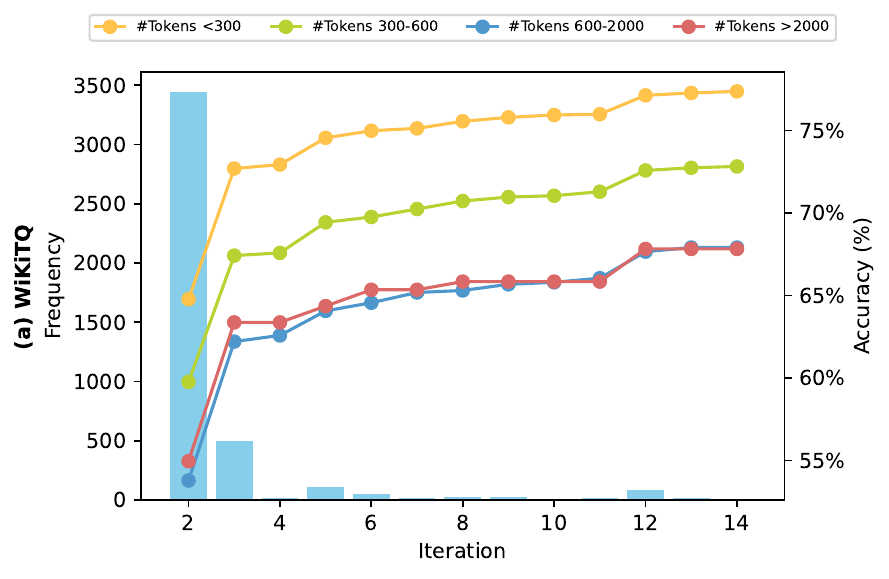}
  % \caption*{(a) \textsc{WikiTQ}}

  % \vspace{0.5em}  

  \includegraphics[width=\columnwidth]{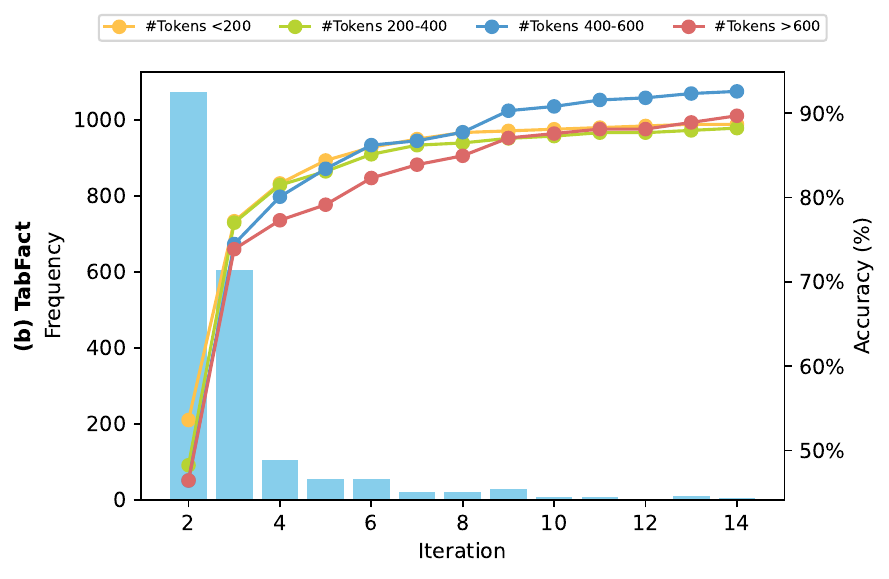}
  % \caption*{(b) \textsc{TabFact}}

  \caption{Analysis of accuracy improvements across reasoning iterations for different table sizes. \textcolor{CornflowerBlue}{Blue} histograms show the distribution of samples by iteration count required for resolution. Line plots track accuracy progression by table size categories.}
  \label{fig:iter_acc}
\end{figure}

\subsection{Impact of Multi-Round Reasoning}
\label{multi_round_appendix}
Figure~\ref{fig:iter_acc} illustrates how accuracy evolves with increasing reasoning iterations across different table sizes. The histograms (blue bars) show the distribution of samples requiring each iteration count, while the line plots track accuracy by table size groups. Due to our framework design, requiring at least one round of Solver and one of Checker—each sample involves a minimum of two LLM calls.
The progressive accuracy improvement pattern is particularly pronounced in \textsc{TabFact}, where initial accuracy starts below 50\% in the first iteration but ultimately surpasses 90\% with sufficient reasoning rounds. This dramatic improvement, nearly doubling accuracy through iterative refinement, demonstrates the substantial limitations of single-pass approaches for fact verification tasks. 

Notably, the benefits of multi-round reasoning vary significantly by table complexity, though with dataset-specific patterns. For \textsc{WikiTQ}, larger tables (>2000 cells) show substantial relative gains, improving by more than 10 percentage points from less than 55\% at first iteration to more than 65\% with extended reasoning, however, their final accuracy still remains below that of smaller tables. In contrast, for \textsc{TabFact}, complex tables (>400 cells, represented by blue and green lines) not only exhibit steeper accuracy growth curves but eventually surpass smaller tables in later iterations. This divergent pattern likely reflects inherent differences in task complexity and dataset characteristics, as \textsc{TabFact} tables are generally smaller (avg \textasciitilde388 tokens) compared to \textsc{WikiTQ} (avg \textasciitilde600 tokens). Nevertheless, both datasets consistently demonstrate that our multi-agent framework provides proportionally greater benefits for complex tables, precisely the scenarios where traditional methods typically struggle most with information overload and reasoning complexity.

These findings strongly support our approach's fundamental premise: while simple cases can be solved with minimal iteration, complex reasoning challenges require structured, iterative refinement through specialized agent collaboration. The early convergence of most samples (approximately 80\% of \textsc{WikiTQ} samples and 70\% of \textsc{TabFact} samples resolved by iteration 3) combined with the continued improvements for complex cases demonstrates both the efficiency and effectiveness of our multi-round approach.

\subsection{Impact of Memory Evolution Strategies}
\begin{table*}[t]
\centering
\resizebox{\textwidth}{!}{%
\begin{tabular}{@{}llccccccccc@{}}
\toprule
\multicolumn{1}{l}{\textbf{Dataset}} & \textbf{\begin{tabular}[c]{@{}c@{}}Evolve Type\end{tabular}} & \textbf{\begin{tabular}[c]{@{}c@{}}Memory \\ Count\end{tabular}} & \textbf{\begin{tabular}[c]{@{}c@{}}Memory \\ Ratio (\%)\end{tabular}} & \textbf{\begin{tabular}[c]{@{}c@{}}Evolution \\ Count\end{tabular}} & \textbf{\begin{tabular}[c]{@{}c@{}}Evolution \\ Ratio (\%)\end{tabular}} & \textbf{\begin{tabular}[c]{@{}c@{}}\# Evolved \\ Memories\end{tabular}} & \textbf{\begin{tabular}[c]{@{}c@{}}Evolution \\ Efficiency ↑\end{tabular}} & \textbf{\begin{tabular}[c]{@{}c@{}}Med. Strengthen \\ Distance\end{tabular}} & \textbf{\begin{tabular}[c]{@{}c@{}}Med. Update \\ Distance\end{tabular}} & \textbf{\begin{tabular}[c]{@{}c@{}}Accuracy\\ (\%) ↑\end{tabular}} \\ \midrule
\multirow{5}{*}{\wikitq} 
 & Always Evolve & 1023 & 23.5\% & 711 & 69.5\% & 936 & 1.32 & 0.64 & 0.64 &  71.42\\
 & Evolve Every $n$ Entries & 1023 & 23.5\% & 147 & 14.4\% & 190 & 1.29 & 0.64 & 0.64 & 69.96 \\
 & LLM-Based & 1023 & 23.5\% & 667 & 65.2\% & 820 & 1.23 & 0.64 & 0.63 & 74.01 \\
 & Never Evolve & 1023 & 23.5\% & 0 & 00.0\% & 0 & 0.00 & NA & NA &  67.89 \\ \midrule
\multirow{5}{*}{\textsc{TabFact}} 
 & Always Evolve & 427 & 21.1\% & 337 & 78.9\% & 423 & 1.26 & 0.61 & 0.61 & 85.84 \\
 & Evolve Every $n$ Entries & 427 & 21.1\% & 68 & 15.9\% & 90 & 1.32 & 0.63 & 0.63 & 85.75 \\
 & LLM-Based & 427 & 21.1\% & 319 & 74.7\% & 372 & 1.17 & 0.60 & 0.58 & 90.66 \\
 & Never Evolve & 427 & 21.1\% & 0 & 00.0\% & 0 & 0.00 & NA & NA & 86.29 \\ \bottomrule
\end{tabular}%
}
\caption{
% Memory system dynamics across different similarity thresholds ($\delta$) on \wikitq~ and \textsc{TabFact} datasets.
Performance of different memory evolution strategies under fixed parameters (similarity threshold $\delta$=0.7, update interval n=5) on the \wikitq~ and \textsc{TabFact} datasets.}
% \vspace{-0.8em}
\label{tab:memory_evolve_type}
\end{table*}
\Cref{tab:memory_evolve_type} compares different memory evolution strategies under fixed parameters ($\delta$=0.7, n=5) on the \wikitq~ and \textsc{TabFact} datasets. The results show that memory evolution has a substantial impact on QA performance. In both datasets, the Never Evolve setting yields the lowest accuracy (67.89\% on \wikitq~ and 86.29\% on \textsc{TabFact}), confirming that static memory limits the system’s adaptability. In contrast, Always Evolve improves performance, but its brute-force nature leads to a large number of updates (69.5\% and 78.9\% evolution ratios on \wikitq~ and \textsc{TabFact}, respectively), introducing redundancy without fully translating into accuracy gains. Similarly, the periodic policy (Evolve Every n Entries) is overly rigid, resulting in moderate efficiency but still lower accuracy than more adaptive approaches.

The LLM-Based strategy consistently achieves the best accuracy (74.01\% on \wikitq~ and 90.66\% on \textsc{TabFact}) while maintaining a balanced evolution efficiency. Unlike naive policies, it selectively evolves memory entries guided by semantic signals, allowing the system to capture meaningful updates while avoiding unnecessary changes. This flexibility explains why its accuracy surpasses both the always-update and fixed-frequency baselines, despite fewer total evolutions. These findings highlight that LLM-guided memory evolution is effective, striking a balance between adaptability and efficiency, and avoiding the rigidity of rule-based strategies.

\section{Example Prompts}
This appendix provides detailed instructions and prompt templates for 4 core agents in our framework: the Solver, the Checker, the Reflector and the Archiver. These agents work collaboratively to tackle table-based question answering tasks through iterative reasoning, verification, and error reflection. 

\begin{figure*}[t]
    \centering
    \includegraphics[width=0.9\textwidth]{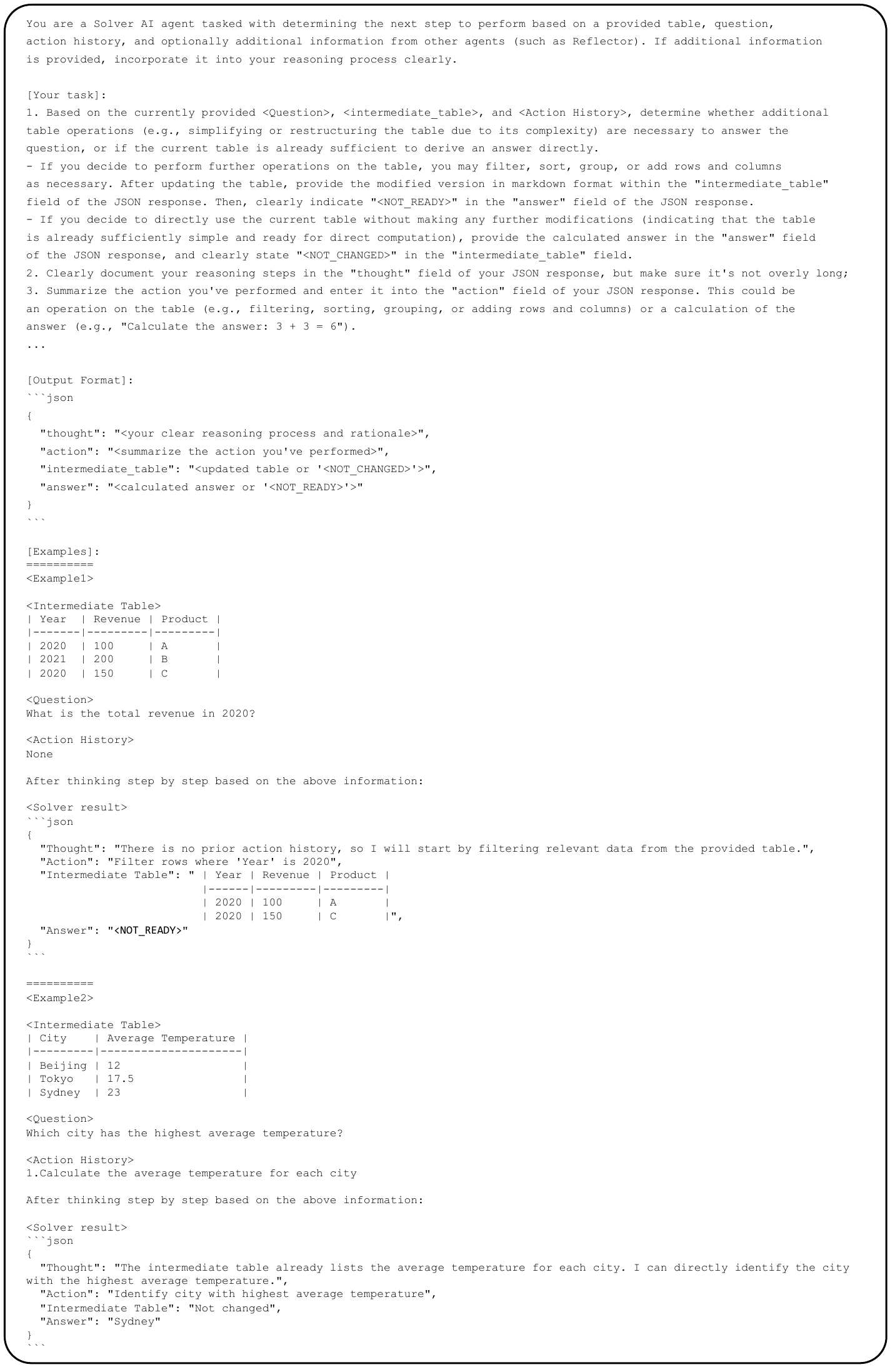}
    \caption{Instructions for the Solver Agent. These instructions guide the agent to perform step-by-step reasoning over the table based on the ReAct paradigm, enabling it to select appropriate operations and generate the final answer.}
    \label{fig:ReasonerExample}
\end{figure*}

Figure~\ref{fig:ReasonerExample} presents the Solver’s step-by-step instructions for interacting with the table based on the ReAct paradigm, including selecting appropriate operations and generating the final answer. 

\begin{figure*}[t]
    \centering
    \includegraphics[width=1\textwidth]{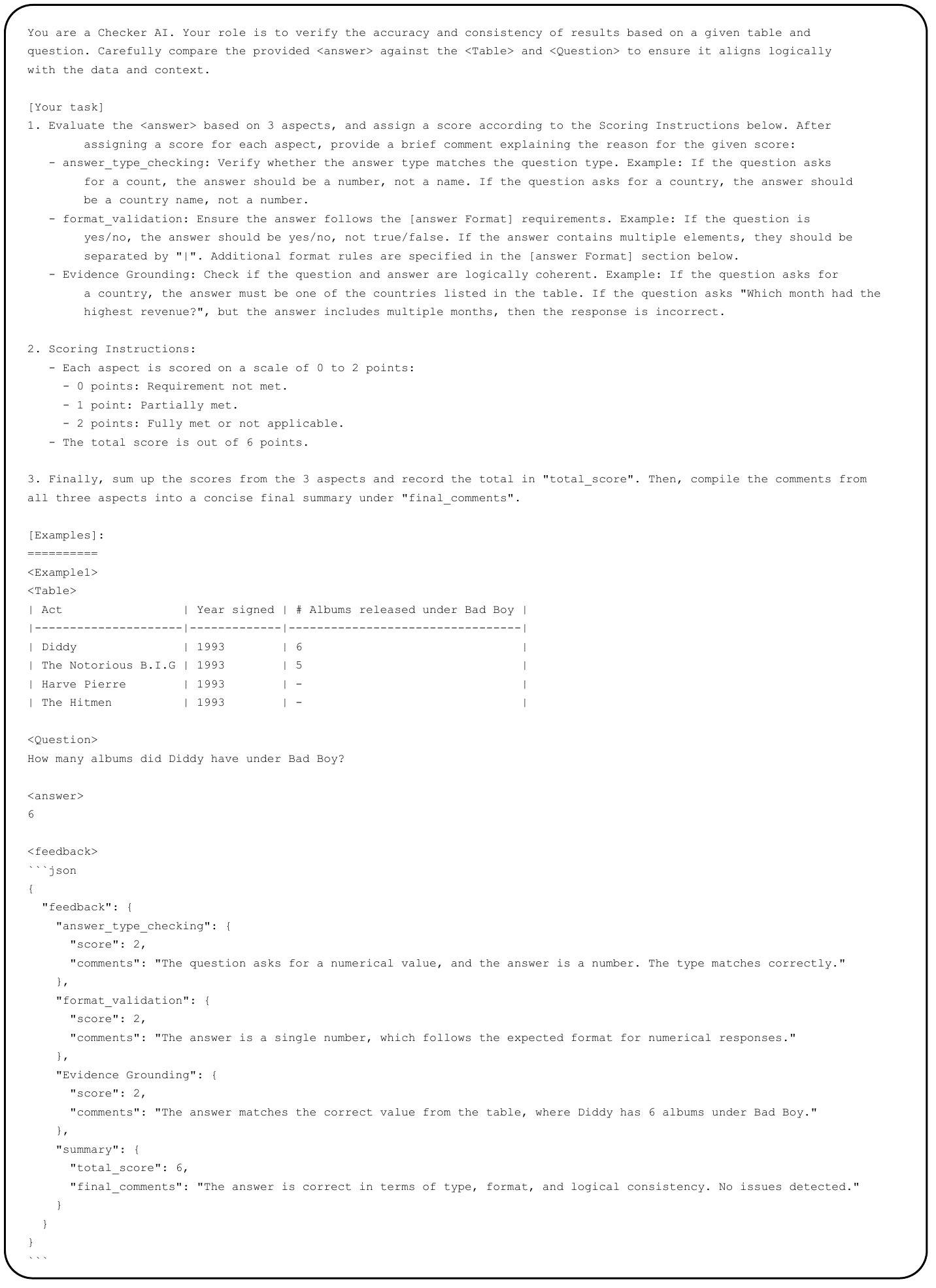}
    \caption{Instructions for the Checker Agent. These instructions guide the agent to evaluate the Reasoner Agent’s answer based on three aspects: answer type, format validation, and evidence grounding.}
    \label{fig:CheckerExample}
\end{figure*}

Figure~\ref{fig:CheckerExample} outlines the Checker Agent’s responsibilities, which involve evaluating the Reasoner’s output from three perspectives: answer type, format validation, and Evidence Grounding. 

\begin{figure*}[t]
    \centering
    \includegraphics[width=0.9\textwidth]{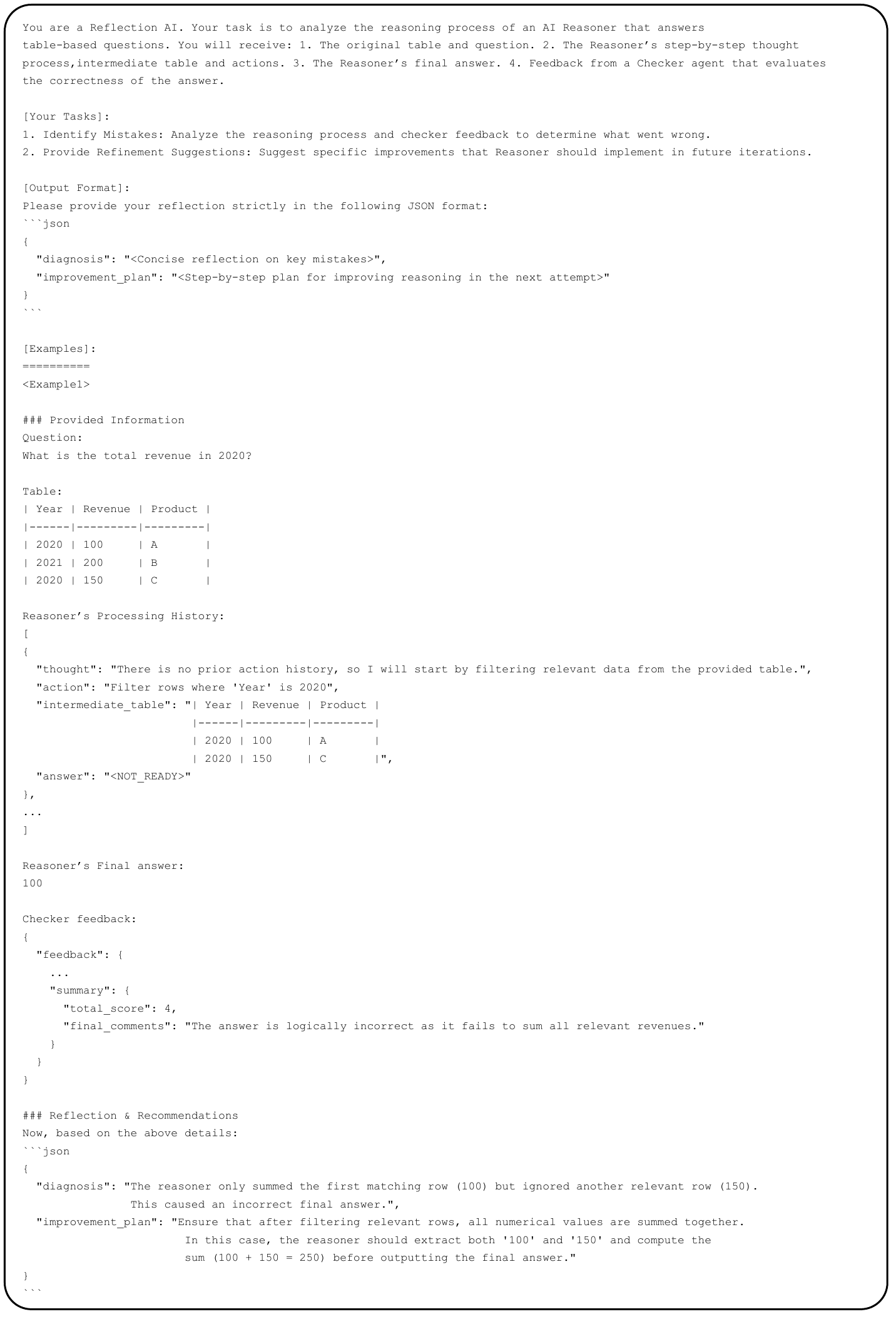}
    \caption{Instructions for the Reflector Agent. These instructions guide the agent to reflect on the provided information—including the table, question, the Reasoner’s answer, and feedback from the Checker Agent—and to identify the cause of the error as well as suggest a direction for improvement.}
    \label{fig:ReflectorExample}
\end{figure*}

Figure~\ref{fig:ReflectorExample} introduces the Reflector Agent, which analyzes feedback from the Checker along with contextual information to identify the source of errors and suggest possible improvements for future reasoning attempts.

\begin{figure*}[t]
    \centering
    \includegraphics[width=0.9\textwidth]{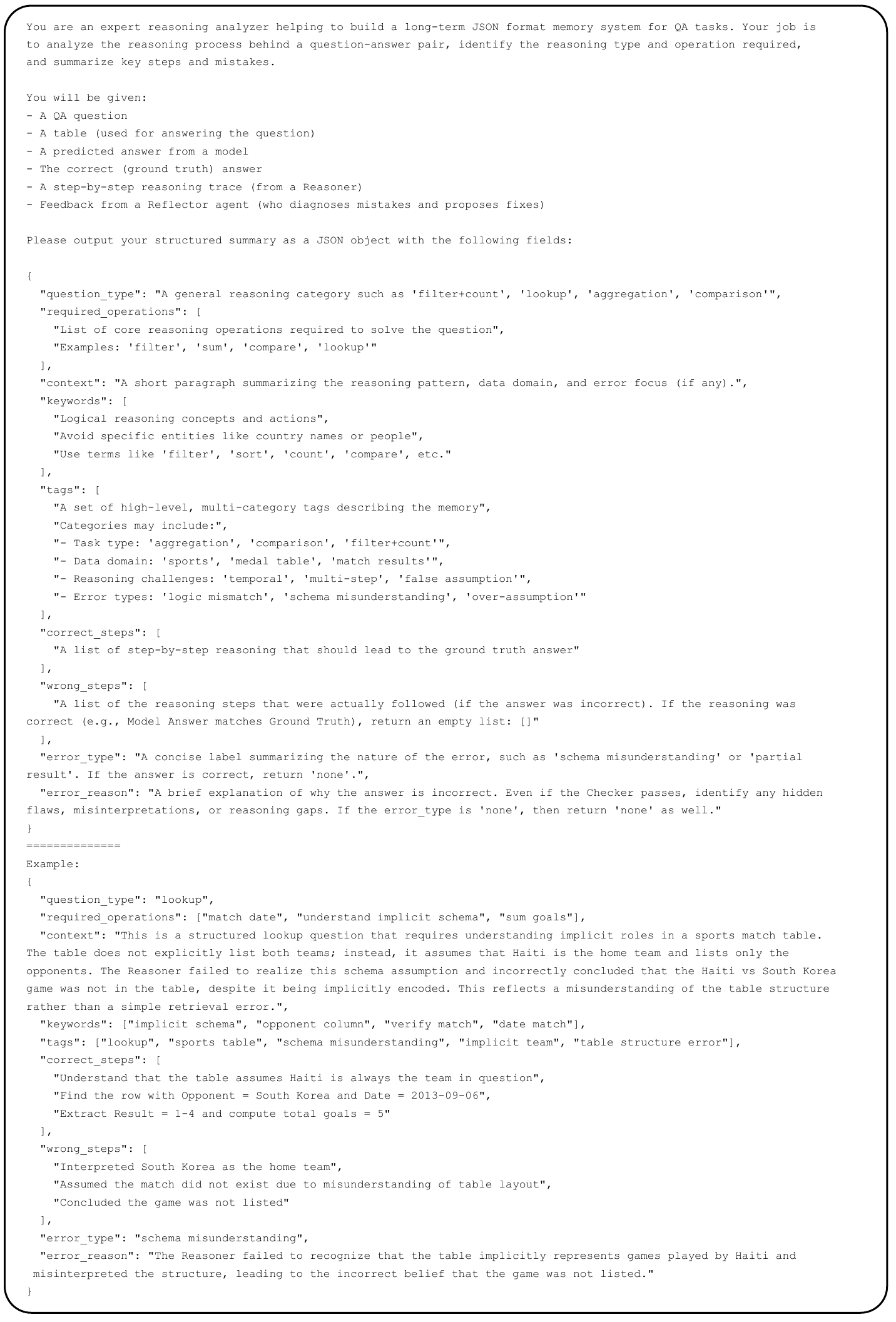}
    \caption{Instructions for the Archiver Agent Memory Summarization Module. These instructions guide the Archiver agent in analyzing the reasoning process of each task, identifying key reasoning types, operations, and errors, and structuring them into a standardized memory note.}
    \label{fig:Advisor_sum}
\end{figure*}

\begin{figure*}[t]
    \centering
    \includegraphics[width=1\textwidth]{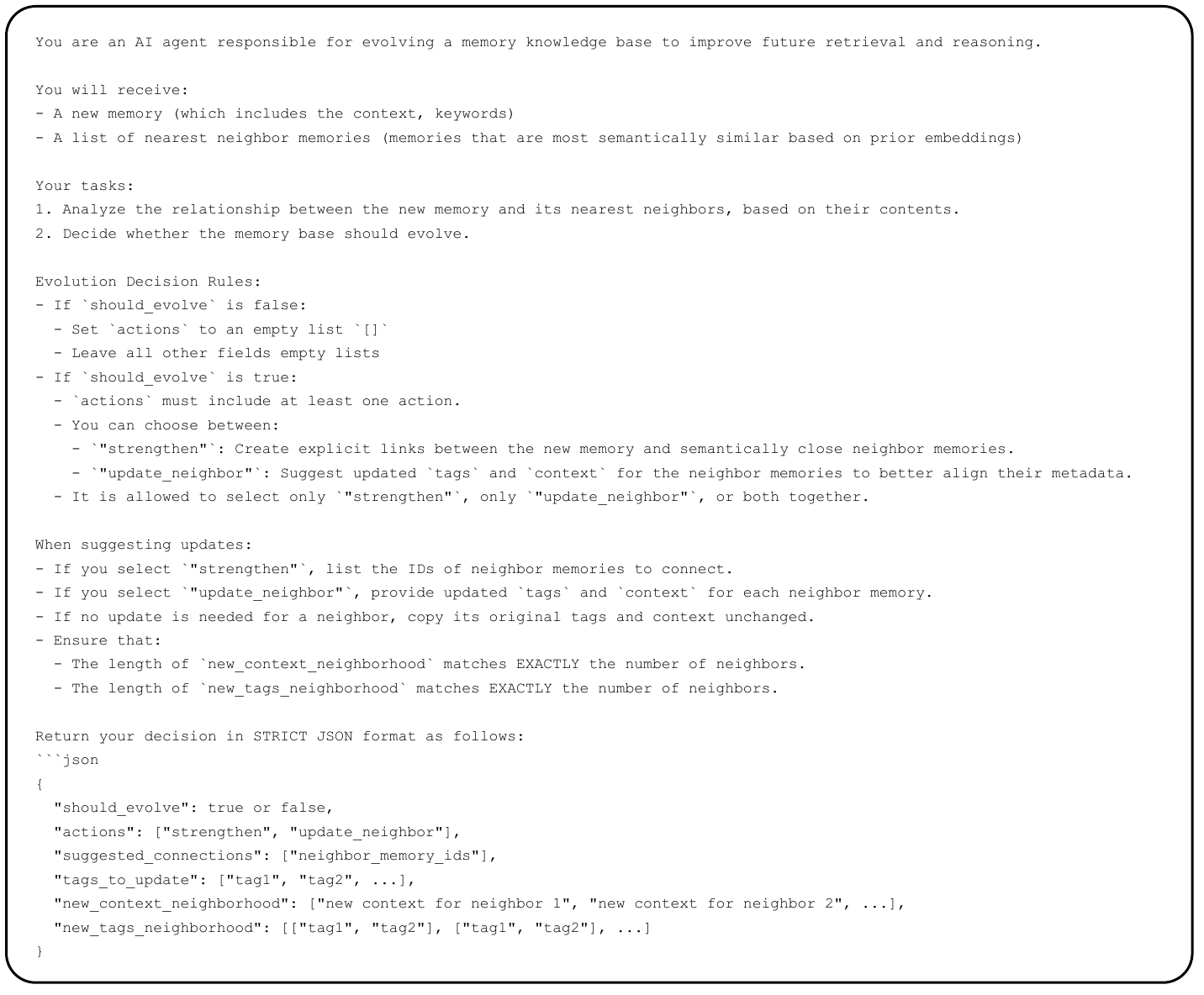}
    \caption{Instructions for the Archiver Agent Memory Evolution Module. These instructions direct the Archiver agent to examine newly created memory notes and their nearest neighbors, determine whether semantic evolution is necessary, and perform actions such as strengthening connections or updating metadata to improve the coherence and retrieval quality of the memory base.}
    \label{fig:Advisor_evol}
\end{figure*}

Finally, Figure~\ref{fig:Advisor_sum} and Figure~\ref{fig:Advisor_evol} detail the Archiver Agent’s dual roles: summarizing each task into structured memory notes and evolving the long-term memory base by refining connections and metadata to enhance future retrieval and reasoning quality.

\label{sec:prompt_examples}


% \textbf{LLM Call Efficiency Across Methods}
% Table~\ref{tab:number_of_calls} compares the number of LLM calls per sample across various methods, based on evaluations on the \textsc{WikiTQ} dataset. \multiagent~ demonstrates a competitive efficiency, requiring only 2–26 calls per question, with an average of 2.56 rounds per sample. This is comparable to Chain-of-Table (1–25 calls), and substantially lower than methods like ReAcTable (15–125 calls) and Dater (100 calls). Importantly, \multiagent~ maintains strong performance while significantly reducing the interaction overhead with the LLM. This efficiency is achieved through its adaptive control over the number of reasoning, checking, and reflecting steps. The reported numbers are based on the configuration where the maximum inner Solver reasoning rounds is set to 5 and the outer Solver retry limit is 4, with similar caps for Checker and Reflector.
% \input{tables/number_of_calls}

\end{document}